\documentclass{article}
\usepackage[utf8]{inputenc}
\usepackage{xcolor}

\newif\ifpreprint

\preprinttrue

\usepackage[preprint]{neurips2022}

\usepackage{subcaption}

\usepackage{amsmath,amsthm,amssymb,amsfonts}
\usepackage{siunitx}
\usepackage{bm}

\usepackage[framemethod=TikZ]{mdframed}
\mdfdefinestyle{MyFrame}{%
    linecolor=black,
    outerlinewidth=.3pt,
    roundcorner=5pt,
    innertopmargin=1pt, %
    innerbottommargin=1pt, %
    innerrightmargin=1pt,
    innerleftmargin=1pt,
    backgroundcolor=black!0!white}

\mdfdefinestyle{MyFrame2}{%
    linecolor=white,
    outerlinewidth=1pt,
    roundcorner=2pt,
    innertopmargin=5pt, %
    innerbottommargin=5pt, %
    innerrightmargin=10pt,
    innerleftmargin=10pt,
    backgroundcolor=black!3!white}

\newcommand{\xhdr}[1]{{\noindent\bfseries #1}.}

\newcommand{\cut}[1]{}

\def\eqref#1{equation~(\ref{#1})}

\def\1{\bm{1}}

\def\vy{{Y}}

\DeclareMathAlphabet{\mathsfit}{\encodingdefault}{\sfdefault}{m}{sl}
\SetMathAlphabet{\mathsfit}{bold}{\encodingdefault}{\sfdefault}{bx}{n}

\def\gE{{\mathcal{E}}}

\def\gI{{\mathcal{I}}}

\def\gL{{\mathcal{L}}}
\def\gM{{\mathcal{M}}}
\def\gN{{\mathcal{N}}}
\def\gO{{\mathcal{O}}}

\def\gT{{\mathcal{T}}}

\def\sH{{\mathbb{H}}}

\def\sP{{\mathbb{P}}}
\def\sQ{{\mathbb{Q}}}

\def\sS{{\mathbb{S}}}
\def\sT{{\mathbb{T}}}

\newcommand{\E}{\mathbb{E}}

\newcommand{\R}{\mathbb{R}}

\newcommand{\Var}{\mathrm{Var}}

\DeclareMathOperator*{\argmin}{arg\,min}

\DeclareMathOperator{\tr}{tr}

\DeclareMathOperator{\supp}{supp}

\newcommand{\partt}[1]{\partial_t #1}
\newcommand{\parts}[1]{\partial_s #1}

\newcommand{\dt}[1]{\frac{\mathrm{d} #1}{\mathrm{d} t}}

\newcommand*{\eg}{{\it e.g.}\@\xspace}
\newcommand*{\ie}{{\it i.e.}\@\xspace}
\newcommand*{\iid}{{\it i.i.d.}\@\xspace}

\def\ds{\,\mathrm{d}s}
\def\dt{\,\mathrm{d}t}

\newcommand{\ddd}{\mathrm{d}}

\newcommand{\gvol}{|G|^{\frac{1}{2}}}
\newcommand{\igvol}{|G|^{-\frac{1}{2}}}
\newcommand{\diverge}{\mathbf{\nabla} \cdot }

\newcommand{\norm}[1]{\left\lVert#1\right\rVert}

\newcommand{\tx}{\tilde{x}}

\newcommand{\overbar}[1]{\mkern 1.5mu\overline{\mkern-1.5mu#1\mkern-1.5mu}\mkern 1.5mu}

\newcommand{\cV}{\mathcal{V}}

\newcommand{\Skew}{\text{Skew}}

\usepackage{booktabs}
\usepackage{enumitem}
\usepackage{algorithm}
\usepackage[noend]{algpseudocode}

\usepackage{url}

\usepackage{thmtools}
\declaretheoremstyle[
bodyfont=\normalfont
]{normalstyle}

\usepackage{thm-restate}

\usepackage[utf8]{inputenc}
\usepackage{pgfplots}
\DeclareUnicodeCharacter{2212}{−}
\usepgfplotslibrary{groupplots,dateplot}
\usetikzlibrary{patterns,shapes.arrows}
\pgfplotsset{compat=newest}
\usepackage{wrapfig}

\usepackage[final]{hyperref} %
\hypersetup{
	colorlinks=true,       %
	linkcolor=blue,        %
	citecolor=blue,        %
	filecolor=magenta,     %
	urlcolor=blue         
}

\usepackage{natbib}
\usepackage{xspace}
\usepackage{soul}
\usepackage{wrapfig}
\usepackage[export]{adjustbox}
\usepackage{cancel}
\usepackage{xcolor}

\usepackage[utf8]{inputenc} %
\usepackage[T1]{fontenc}    %

\title{Riemannian Diffusion Models}

\author{%
  Chin-Wei Huang$^*$,\,\, Milad Aghajohari$^*$,\,\, Avishek Joey Bose\\ \textbf{Prakash Panangaden,\,\, Aaron Courville }\\
  University of Montreal \& McGill University \& Mila \\
  \texttt{\{chin-wei.huang, milad.aghajohari, aaron.courville\}@umontreal.ca} \\
  \texttt{joey.bose@mail.mcgill.ca, prakash@cs.mcgill.ca} \\
}

\begin{document}

\maketitle

\begin{abstract}
    Diffusion models are recent state-of-the-art methods for image generation and likelihood estimation.
    In this work, we generalize continuous-time diffusion models to arbitrary Riemannian manifolds and derive a variational framework for likelihood estimation. %
    Computationally, we propose new methods for computing the Riemannian divergence which is needed in the likelihood estimation.  
    Moreover, in generalizing the Euclidean case, we prove that maximizing this variational lower-bound is equivalent to Riemannian score matching.
    Empirically, we demonstrate the expressive power of Riemannian diffusion models on a wide spectrum of smooth manifolds, such as spheres, tori, hyperboloids, and orthogonal groups. 
    Our proposed method achieves new state-of-the-art likelihoods on all benchmarks. %

\end{abstract}

\section{Introduction}
\label{sec:intro}
\everypar{\looseness=-1}
By learning to transmute noise, generative models seek to uncover the underlying generative factors that give rise to observed data. 
These factors can often be cast as inherently geometric quantities as the data itself need not lie on a flat Euclidean space. 
Indeed, in many scientific domains such as 
high-energy physics \citep{brehmer2020flows}, directional statistics \citep{mardia2009directional}, geoscience \citep{mathieu2020riemannian}, computer graphics \citep{kazhdan2006poisson}, and linear biopolymer modeling such as protein and RNA \citep{mardia2008multivariate,boomsma2008generative, frellsen2009probabilistic}, data is best represented on a Riemannian manifold with a \textit{non-zero curvature}. 
Naturally, to effectively capture the generative factors of these data, we must take into account the geometry of the space when designing a learning framework.   

Recently, diffusion based generative models have emerged as an attractive model class that not only achieve likelihoods comparable to state-of-the-art autogressive models \citep{kingma2021variational} but match the sample quality of GANs without the pains of adversarial optimization \citep{dhariwal2021diffusion}. 
Succinctly, a diffusion model consists of a fixed %
Markov chain that progressively transforms data to a prior defined by the inference path, and a generative model which is another Markov chain that is learned to invert the inference process \citep{ho2020denoising, song2021score}. 

While conceptually simple, the learning framework can have a variety of perspectives and goals. 
For example, \citet{huang2021variational} provide a variational framework for general continuous-time diffusion processes on Euclidean manifolds as well as a functional Evidence Lower Bound (ELBO) that can be equivalently shown to be minimizing an implicit score matching objective. 
At present, however, much of the success of diffusion based generative models and its accompanying variational framework is purpose built for Euclidean spaces, and more specifically, image data. It does not easily translate to general Riemannian manifolds.

In this paper, we introduce Riemannian Diffusion Models (RDM)---generalizing conventional diffusion models on Euclidean spaces to arbitrary Riemannian manifolds.
Departing from diffusion models on Euclidean spaces, our approach uses the Stratonovich SDE formulation for which the conventional chain rule of calculus holds, which, as we demonstrate in section~\S\ref{sec:rdm}, can be exploited to define diffusion on a Riemannian manifold. 
Furthermore, we take an extrinsic view of geometry by defining the Riemannian manifold of interest as an embedded sub-manifold within a higher dimensional (Euclidean) ambient space. 
Such a choice enables us to define both our inference and generative SDEs using the coordinate system of the ambient space, greatly simplifying the implementation of the theory developed using the intrinsic view. %

\xhdr{Main Contributions} We summarize our main contributions below:
\begin{itemize}[noitemsep,topsep=0pt,parsep=0pt,partopsep=0pt,label={\large\textbullet},leftmargin=*]
    \item  We introduce a variational framework %
    built on the Riemannian Feynman-Kac representation 
    and
    Giransov's theorem.
    In Theorem \ref{thm:ct-elbo} we derive a Riemannian continuous-time ELBO, strictly generalizing the CT-ELBO in \citet{huang2021variational} and prove in Theorem \ref{thm:score} that its maximization is equivalent to Riemannian score matching for marginally equivalent SDEs (Theorem \ref{thm:equivalent}).
    \item To compute the Riemannian CT-ELBO it is necessary to compute the Riemannian divergence of our parametrized vector field, for which we introduce a QR-decomposition-based method that is computationally efficient for low dimensional manifolds as well a projected Hutchinson method for scalable unbiased estimation. Notably, our approach does not depend on the closest point projection which may not be freely available for many Riemannian manifolds of interest.
    \item We also provide a variance reduction technique to estimate the Riemannian CT-ELBO objective that leverages importance sampling with respect to the time integral, which crucially avoids carefully designing the noise schedule of the inference process. 
    \item Empirically, we validate our proposed models on spherical manifolds towards modelling natural disasters as found in earth science datasets, products of spherical manifolds (tori) for protein and RNA, synthetic densities on hyperbolic spaces and orthogonal groups. 
    Our empirical results demonstrate that RDM leads to new state-of-art likelihoods over prior manifold generative models.%
\end{itemize}

\cut{

Our key contributions are summarized as follows: 
\begin{itemize}[noitemsep,topsep=0pt,parsep=0pt,partopsep=0pt,label={\large\textbullet},leftmargin=*]
\item We extend the diffusion framework from Euclidean spaces to arbitrary Riemannian manifolds.
\item We derive a Riemannian ELBO for continuous-time diffusion models in Thm. \ref{thm:ct-elbo} by leveraging the Riemannian Feynmann-Kac representation and Girsanov's theorem.
\item We provide a computationally efficient time-reversed fixed-inference parametrization of the SDE in the ambient space leading to significantly more stable and efficient training.%
\item We connect Riemannian score matching within our variational maximum likelihood framework, harmonizing two separate frameworks in a unified view.
\item We empirically demonstrate the expressive power of Riemannian diffusion models on spherical, toroidal, and hyperbolic manifolds and achieve state-of-the-art log likelihoods and sample quality.

\end{itemize}

1:
SOTA quality and likelihood.

2:
Motivation. 
Biology.
Physics.
Chemistry. 
Computer graphics.

3: 
Contribution.

\begin{itemize}
    \item Theoretical contribution 
\begin{itemize}
    \item extending the diffusion framework on $\R^d$,
    \begin{itemize}
        \item Deriving a marginal likelihood using (Riemannian) Feynman-Kac
        \item Deriving a Riemannian ELBO for continuous-time diffusion model. Theorem~\ref{thm:ct-elbo}
        \item TIme-reversed parameterization (computational efficiency). Section~\ref{sec:fixed-inf}
    \end{itemize}
    \item Connect riemannian score matching with variational maximum likelihood.
    Section~\ref{sec:score}
\end{itemize}

\item Part of the theory leads to computational benefits: 

\begin{itemize}
    \item divergence identity, QR trick more universal (for lower dimensional problem).
    Section~\ref{sec:div}.
    \item Variance reduction. Section~\ref{sec:vr}
\end{itemize}

\item Diffusion models work on general Riemannian manifold.
Performance. SOTA (If we have it).

\end{itemize}
}

\section{Background}
\label{sec:background}
\everypar{\looseness=-1}
In this section, we provide the necessary background on diffusion models and key concepts
from Riemannian geometry that we utilize to build RDMs.  For a
short review of the latter, see Appendix~\ref{app:riem} or \citet{Ratcliffe94} for a more
comprehensive treatment of the subject matter.

\subsection{Euclidean diffusion models}
A diffusion model can be defined as the solution to the (It\^o) SDE \citep{oksendal2003stochastic},
\begin{align}
    \ddd X = \mu \dt + \sigma \,\ddd B_t,
    \label{eq:genmodel-euclidean}
\end{align}
with the initial condition $X_0$ following some unstructured prior $p_0$ such as the standard normal distribution, where $B_t$ is a standard Brownian motion, and $\mu$ and $\sigma$ are the drift and diffusion coefficients of the diffusion process,
which control the deterministic forces driving the evolution and the amount of noise injected at each time step.
This provides us a way to sample from the model, via numerically solving the dynamics from $t=0$ to $t=T$ for some fixed termination time $T$. 
To train the model via maximum likelihood, we require an expression for the log marginal density of $X_T$, denoted by $\log p(x, T)$, which is generally intractable.

The marginal likelihood can be represented using a stochastic instantaneous change-of-variable formula, by applying the Feynman-Kac theorem to the Fokker-Planck PDE of the density. 
An application of Girsanov's theorem followed by an application of Jensen's inequality leads to the following variational lower bound~\citep{huang2021variational, song2021maximum}:
\begin{align}
    \log p(x,T) \geq \E\left[\log p_0(Y_T) - \int_0^T \left( \frac{1}{2}\norm{a(Y_s, s)}_2^2 + \nabla\cdot \mu(Y_s, T-s) \right) \ds  \, \Bigg\vert \, \vy_0=x  \right]
\end{align}
where $a$ is the variational degree of freedom, $\nabla\cdot$ denotes the (Euclidean) divergence operator, and $Y_s$ follows the inference SDE (the generative coefficients are evaluated in reversed time, \ie $T-s$)
\begin{align}
    \ddd Y = (-\mu +\sigma a)\ds +\sigma \, \ddd \hat{B}_s
\end{align}
with $\hat{B}_s$ being another Brownian motion. 
This is known as the continuous-time evidence lower bound, or the CT-ELBO for short.

\subsection{Riemannian manifolds}
\label{sec:riemannian-manifolds}
We work with a $d$-dimensional Riemannian manifold $(\gM, g)$ embedded in a higher dimensional ambient space $\R^m$, for $m>d$. This assumption does not come with a loss of generality, since any Riemannian manifold can be isometrically embedded into a Euclidean space by the \emph{Nash embedding theorem} \citep{gunther1991isometric}.
In this case, the metric $g$ coincides with the pullback of the Euclidean metric by the inclusion map. Now, given a coordinate chart $\varphi: \mathcal{M} \to \mathbb{R}^d$ and its inverse $\psi=\varphi^{-1}$, we can define  $\tilde{E}_j$ for $j=1,\cdots,d$ to be the basis vectors of the tangent space $\gT_x\gM$ at point $x \in \mathcal{M}$. The tangent space can be understood as the pushforward of the Euclidean derivation of the patch space along $\psi$; \ie, for any smooth function $f\in C^\infty(\gM)$, 
$\tilde{E}_j(f) = \frac{\partial}{\partial \tx_j} f\circ \psi$.

We denote by $P_x$ the orthogonal projection onto the linear subspace spanned by the column vectors of the Jacobian $J_x = \ddd \psi/\ddd \tx$. 
Specifically, $P_x$ can be constructed via $P_x = J_x(J_x^TJ_x)^{-1}J_x^T$. Note that this subspace is isomorphic to the tangent space $\gT_x\gM$, which itself is a subspace of $\gT_x\R^m$.
As a result, we identify this subspace with $\gT_x\gM$.
Lastly, we refer to the action of $P_x$ as the projection onto the tangential subspace, and $P_x$ itself as the tangential projection.

\subsection{SDE on manifolds}
\label{sec:sde-manifolds}
Unlike Euclidean spaces, Riemannian manifolds generally do not possess a vector space structure.
This prevents the direct application of the usual (stochastic) calculus. 
We can resolve this by defining the process via test functions. Specifically, 
let $V_k$ be a family of smooth vector fields on $\gM$, and let $Z^k$ be a family of semimartingales \citep{protter_stochastic_2005}. 
Symbolically, we write
\begin{align}
    \ddd X_t = \sum_{k} V_k(X_t) \circ \ddd Z_t^k
    \quad \textnormal{ if }\quad \ddd f(X_t) = \sum_{k} V_k(f)(X_t) \circ \ddd Z_t^k
    \label{eq:sde-abstract}
\end{align}
for any $f\in C^\infty(\gM)$ \citep{hsu2002stochastic}. 
The $\circ$ in the second differential equation is to be interpreted in the Stratonovich sense \citep{protter_stochastic_2005}.
The use of the Stratonovich integral is the first step deviating from the Euclidean diffusion model (\ref{eq:genmodel-euclidean}), as the It\^o integral does not follow the usual chain rule.

Working with this abstract definition is not always convenient, so instead we work with specific coordinates of $\gM$. 
Let $\varphi$ be a chart, and let $\tilde{v}= (\tilde{v}_{jk})$ be a matrix representing the coefficients of $V_k$ in the coordinate basis---i.e.
$V_k(f) = \sum_{j=1}^d \tilde{v}_{jk} \frac{\partial }{\partial \tx_j} f\circ\varphi^{-1}\big|_{\tx=\varphi(x)}$.
This allows us to write %
$\ddd \varphi(X_t) = \tilde{v} \circ \ddd Z$. Similarly, suppose $\gM$ is a submanifold embedded in $\R^m$, and denote by $v=(v_{ik})$ the coefficients wrt the Euclidean basis.
$v$ and $\tilde{v}$ are related by $v =\frac{\ddd \varphi^{-1}}{\ddd\tx} \tilde{v}$. 
Then we can express the dynamics of $X$ as a regular SDE using the Euclidean space's coefficients $\ddd X = v \circ \ddd Z$. Notably, by the relation between $v$ and $\tilde{v}$, the column vectors of $v$ are required to lie in the span of the column vectors of the Jacobian $\frac{\ddd \varphi^{-1}}{\ddd\tx}$ which restricts the dynamics to move tangentially on $\gM$.

\section{Riemannian diffusion models}
\label{sec:rdm}
We now develop a variational framework to estimate the likelihood of a diffusion model defined on a Riemannian manifold $(\gM, g)$. 
Let $X_t \in \gM$ be a process solving the following SDE:
\begin{align}
    \textnormal{Generative SDE:} \qquad \ddd X = V_0 \dt + V \circ \ddd B_t, \qquad X_0 \sim p_0
    \label{eq:genmodel}
\end{align}
where $V_0$ and 
the columns of the diffusion matrix\footnote{The multiplication is interpreted similarly to matrix-vector multiplication, \ie $V\circ\ddd B_t = \sum_{k=1}^w V_k \circ \ddd B^k_t$.} $V:=[V_1,\cdots, V_w]$ are smooth vector fields on $\gM$, and $B_t$ is a $w$-dimensional Brownian motion.  
The law of the random variable $X_t$ can be written as $p(x, t) \,\mu(\ddd x)$, where $p(x, t)$ is the probability density function and $\mu$ is the $d$-dimensional Hausdorff measure on the manifold associated with the Riemannian volume density. 
Let $V\cdot\nabla$ be a differential operator defined by $(V\cdot \nabla_g) U := \sum_{k=1}^w (\nabla_g \cdot U_k) V_k$, where $\nabla_g \cdot U_k$ denotes the \emph{Riemannian divergence} of the vector field $U_k$:
\begin{align}
    \nabla_g \cdot U_k  = \igvol
    \sum_{j=1}^d\frac{\partial}{\partial \tx_j}(\gvol \tilde{u}_{jk} ).
    \label{eq:riem-div-coord}
\end{align}
Our first result is a stochastic instantaneous change-of-variable formula for the Riemannian SDE by applying the Feynman-Kac theorem to the Fokker Planck PDE of the density $p(x,t)$.
\begin{mdframed}[style=MyFrame2]
\begin{restatable}[\textbf{Marginal Density}]{thm}{marginal}
\label{thm:marginal-density}
The density $p(x,t)$ of the SDE (\ref{eq:genmodel}) can be written as
\begin{equation}
p(x, t)=\mathbb{E}\left[ p_0\left(Y_{t}\right) \exp \left(- \int_{0}^{t} 
\nabla_g\cdot \left(V_0 - \frac{1}{2}  (V \cdot \nabla_g) V\right)
d s\right) \, \Bigg\vert \, \vy_0=x \right]
\end{equation}
where the expectation is taken wrt the following process induced by a Brownian motion $B_s'$
\begin{align}
    \ddd Y= ( -V_0 + (V\cdot \nabla_g)  V) \ds+ V \circ \ddd B_s'.
    \label{eq:fk-sde}
\end{align}
\end{restatable}
\end{mdframed}

For effective likelihood maximization, we require access to $\log p$ and its gradient. Towards this goal, we prove the following Riemannian CT-ELBO which serves as our training objective and follows from an application of change of measure (Girsanov's theorem) and Jensen's inequality.
\begin{mdframed}[style=MyFrame2]
\begin{restatable}[\textbf{Riemannian CT-ELBO}]{thm}{rctelbo}
\label{thm:ct-elbo}
Let $\hat{B}_s$ be a $w$-dimensional Brownian motion, 
and let $Y_s$ be a process solving the following
\begin{align}
    \textnormal{Inference SDE:} \qquad \ddd Y = \left(- V_0 + (V\cdot \nabla_g) V + {Va} \right) \ds + V \circ \ddd \hat{B}_s,
    \label{eq:inference-sde}
\end{align}
where $a:\R^m\times[0,T]\rightarrow \R^m$ is the variational degree of freedom. 
Then we have
\begin{align}
    \log p(x,T) \geq %
    \E\left[ 
    \log p_0(\vy_T) 
    -  \int_0^T \frac{1}{2} \norm{a(Y_s, s)}_2^2 + \nabla_g\cdot \left(V_0 - \frac{1}{2}  (V\cdot\nabla_g)V \right)\ds
    \, \Bigg\vert \, \vy_0=x
    \right],
    \label{eq:ct-elbo}
\end{align}
where all the generative degree of freedoms $V_k$ are evaluated in the reversed time direction.%
\end{restatable}
\end{mdframed}

\subsection{Computing Riemannian divergence}
\label{sec:div}

Similar to the Euclidean case, computing the Riemannian CT-ELBO requires computing the divergence ``$\nabla_g \cdot$'' of a vector field\cut{ to keep track of the probability flux and maintain the conservation of mass.}, which can be achieved by applying the following identity.%

\begin{mdframed}[style=MyFrame2]
\begin{restatable}[\textbf{Riemannian divergence identity}]{prop}{divergence}
Let $(M, g)$ be a $d$-dimensional Riemannian manifold. 
For any smooth vector field $V_k \in \mathfrak{X}(\gM)$, the following identity holds:
\begin{align}
    \nabla_g\cdot V_k %
    = \sum_{j=1}^d  \left\langle\nabla_{\tilde{E}_j} V_k, \tilde{E}^j \right\rangle_g.
    \label{eq:riem-div-intrinsic}
\end{align}
Furthermore, if the manifold is a submanifold embedded in the ambient space $\R^m$ equipped with the induced metric $g=\iota^*\bar{g}$, then
\begin{align}
    (\nabla_g\cdot V_k)(x) = \tr\left(P_x \frac{\ddd {v_k}}{\ddd x} P_x\right),
    \label{eq:riem-div-ambient}
\end{align}
where $v_k=(v_{1k},\cdots,v_{mk})$ are the ambient space coefficients $V_k=\sum_{i=1}^m {v}_{ik} \frac{\partial}{\partial x_i}$ and $P_x$ is the orthogonal projection onto the tangent space.
\end{restatable}
\end{mdframed}

\xhdr{Intrinsic coordinates} The patch-space formula (\ref{eq:riem-div-coord}) can be used to compute the Riemannian divergence. 
This view was adopted by \citet{mathieu2020riemannian}, where they combined the Hutchinson trace identity and the internal coordinate formula to estimate the divergence.
The drawbacks of this framework include: (1) obtaining local coordinates may be difficult for some manifolds, hindering generality in practice; (2) we might need to change patches, which complicates implementations; and (3) the inverse scaling of $\sqrt{|G|}$ might result in numerical instability and high variance.

\xhdr{Closest-point projection} The coordinate-free expression (\ref{eq:riem-div-intrinsic}) leads to the closest-point projection method proposed by \citet{rozen2021moser}. 
Concretely, define the closest-point projection by $\pi(x) := \argmin_{y \in \mathcal{M}} \norm{x - y}$, where $\norm{\cdot}$ is the Euclidean norm. 
Let $V_k(x)$ be the derivation corresponding to the ambient space vector $v_k(x) = P_{\pi(x)} u(\pi(x))$ for some unconstrainted $u:\R^m \rightarrow\R^m$. 
\citet{rozen2021moser} showed that $\nabla_g \cdot V_k(x) =\nabla \cdot v_k(x)$, since $v_k$ is infinitesimally constant in the normal direction to $\gT_x \mathcal{M}$.
This allows us to compute the divergence directly in the ambient space.
However, the closest-point projection map $\pi$ may not always be easily obtained.

\xhdr{QR decomposition} 
An alternative to the closest-point projection is to instead search for an orthogonal basis for $\gT_x \mathcal{M}$. 
Let $Q = [e_1, \cdots, e_d, n_1, \cdots, n_{m-d}]$ be an orthogonal matrix whose first $d$ columns span the $\gT_x\gM$, and the remaining $m-d$ vectors span its orthogonal complement $\gT_x\gM^{\perp}$. 
To construct $Q$ we can simply sample $d$ vectors---\eg from $\gN(0,1)$--in the ambient space and orthogonally project them to $\gT_x \mathcal{M}$ using $P_x$. 
These vectors, although not orthogonal yet, form a basis for $\gT_x \mathcal{M}$. 
Next we concatenate them with $m-d$ random vectors and apply a simple $QR$ decomposition to retrieve an orthogonal basis.
Using $Q$ we may rewrite \eqref{eq:riem-div-ambient} as follows:
\begin{align}
    (\nabla_g\cdot V_k)(x) 
    = \tr\left(Q Q^\top P_x \frac{\ddd {v_k}}{\ddd x} P_x\right)
    = \tr\left((P_xQ)^\top \frac{\ddd {v_k}}{\ddd x} P_xQ\right)
    = \sum_{j=1}^d e_j^\top \frac{\ddd {v_k}}{\ddd x} e_j
\end{align}
where we used (1) the orthogonality of $Q$, %
(2) the cyclic property of trace, (3) and the fact that $P_xe_j=e_j$ and $P_x n_j = 0$. %
In practice, concatenation with the remaining $m-d$ vectors is not needed as they are effectively not used in computing the divergence, speeding up computation when $m\gg d$. 
Moreover, the vector-Jacobian product can be computed in $\gO(m)$ time using reverse-mode autograd and importantly does not require the closest-point projection $\pi$.

\xhdr{Projected Hutchinson}
When QR is too expensive for 
higher dimensional problems, the Hutchinson trace estimator \citep{hutchinson1989stochastic} can be employed within the extrinsic view representation (\ref{eq:riem-div-ambient}).
For example, let $z$ be a standard normal vector (or a Rademacher vector), we have $(\nabla_g\cdot V_k)(x) = \E_{z\sim \gN, z' = P_xz} [z'^\top \frac{\ddd v_k}{\ddd x} z']$.
Different from a direct application of the trace estimator to the closest-point method, we directly project the random vector to the tangent subspace.
Therefore, the closest-point projection is again not needed.

\subsection{Fixed-inference parameterization}
\label{sec:fixed-inf}

Following prior work \citep{sohl2015deep, ho2020denoising, huang2021variational}, we let the inference SDE (\ref{eq:inference-sde}) be defined as a simple noise process taking observed data to unstructured noise:
\begin{align}
    \ddd Y = U_0 \dt + V \circ \ddd \hat{B}_s,
    \label{eq:fixed-inference}
\end{align}
where $U_0=\frac{1}{2}\nabla_g \log p_0$ and $V$ is the tangential projection matrix; that is, $V_k(f)(x) = \sum_{j=1}^m (P_x)_{jk} \frac{\partial f}{\partial x_j}$ for any smooth function $f$. 
This is known as the \emph{Riemannian Langevin diffusion} \citep{girolami2011riemann}. 
As long as $p_0$ satisfies a log-Sobolev inequality, the marginal distribution of $Y_s$ (\ie the aggregated posterior) converges to $p_0$ at a linear rate in the KL divergence \citep{wang2020fast}. 
For compact manifolds, we set $p_0$ to be the uniform density, which means $U_0=0$, and (\ref{eq:fixed-inference}) is reduced to the extrinsic construction of Brownian motion on $\gM$ \citep[Section 1.2]{hsu2002stochastic}. 
The benefits of this fixed-inference parameterization are the following:

\xhdr{Stable and Efficient Training} 
With the fixed-inference parameterization we do not need to optimize the vector fields that generate $Y_s$, and the Riemannian CT-ELBO can be rewritten as:
\begin{align}
    \E[\log p_0(Y_T)] -  \int_0^T \E_{Y_s}\left[\,
    \frac{1}{2} \norm{a(Y_s, s)}_2^2 + \nabla_g\cdot \left(V_0 - \frac{1}{2}  (V\cdot\nabla_g)V \right)
    \, \Bigg\vert \, Y_0=x\right] \ds,
    \label{eq:ct-elbo-time}
\end{align}
where the first term is a constant wrt the model parameters (or it can be optimized separately if we want to refine the prior), and the time integral of the second term can be estimated via importance sampling (see Section~\ref{sec:vr}). 
A sample of $Y_s$ can be drawn cheaply by numerically integrating (\ref{eq:fixed-inference}), without requiring a stringent error tolerance (see Section~\ref{sec:tori} for an empirical analysis),
which allows us to estimate the time integral in (\ref{eq:ct-elbo-time}) by evaluating $a(Y_s,s)$ at a single time step $s$ only.

\xhdr{Simplified Riemannian CT-ELBO}
The CT-ELBO can be simplified as the differential operator $V \cdot \nabla_g$ applied to $V$ yields a zero vector when $V$ is the tangential projection. 
\begin{mdframed}[style=MyFrame2]
\begin{restatable}{prop}{projection}
\label{prop:proj}
If $V$ is the tangential projection matrix, then $(V\cdot\nabla_g)V = 0$.
\end{restatable}
\end{mdframed}
This means that we can express the generative SDE $V_0$ using the variational parameter $a$ via
\begin{align}
    \ddd X = (V a(X, T-t) - U_0(X, T-t) ) \dt + V \circ \ddd \hat{B}_t,
    \label{eq:fixed-inference-gen}
\end{align}
with the corresponding Riemannian CT-ELBO:
\begin{align}
    \E[\log p_0(Y_T)] -  \int_0^T \E_{Y_s}
    \left[ \frac{1}{2} \norm{a}^2_2 + \nabla_g\cdot (V a- U_0) \, \Bigg\vert \,  Y_0=x\right] \ds.
    \label{eq:ct-elbo-time2}
\end{align}

\subsection{Variance reduction}
\label{sec:vr}
The inference process can be more generally defined to account for a time reparameterization. 
In fact, this leads to an equivalent model
if one can find an invariant representation of the temporal variable.
Learning this time rescaling can help to reduce variance \citep{kingma2021variational}.

In principle, we can adopt the same methodology, but this would further complicate the parameterization of the model. 
Alternatively, we opt for a simpler view for variance reduction via importance sampling.
We estimate the time integral ``$\int \dots \ds$'' in (\ref{eq:ct-elbo-time2}) using the following estimator:
\begin{align}
    \gI := \frac{1}{q(s)}\left(\frac{1}{2} \norm{a}^2_2 + \nabla_g\cdot (Va-U_0)\right) \qquad \textnormal{where } s\sim q(s) \textnormal{ and } Y_s\sim q(Y_s \mid Y_0),
\end{align}
where $q(s)$ is a proposal density supported on $[0,T]$. 
We parameterize $q(s)$ using a 1D monotone flow~\citep{huang2018neural}.
As the expected value of this estimator is the same as the time integral in (\ref{eq:ct-elbo-time2}), it is unbiased.
However, this means we cannot train the proposal distribution $q(s)$ by maximizing this objective, since the gradient wrt the parameters of $q(s)$ is zero in expectation. 
Instead, we minimize the variance of the estimator by following the stochastic gradient wrt $q(s)$
\begin{align}
    \nabla_{q(s)} \Var(\gI) = \nabla_{q(s)} \E[\gI^2] - \cancel{\nabla_{q(s)} \E[\gI]^2} = \nabla_{q(s)} \E[\gI^2]. 
\end{align}
The latter can be optimized using the reparameterization trick~\citep{kingma2014auto} and is a well-known variance reduction method in a multitude of settings  \citep{luo2020sumo,tucker2017rebar}.
It can be seen as minimizing the $\chi^2$-divergence from a density proportional to the magnitude of $\E_{Y_s}[\gI]$ \citep{dieng2017variational,muller2019neural}.

\subsection{Connection to score matching}
\label{sec:score}
In the Euclidean case, it can be shown that maximizing the variational lower bound of the fixed-inference diffusion model (\ref{eq:fixed-inference-gen}) is equivalent to score matching \citep{ho2020denoising, huang2021variational, song2021maximum}. 
In this section, we extend this connection to its Riemannian counterpart. 

Let $q(y_s, s)$ be the density of $Y_s$ following (\ref{eq:fixed-inference}), marginalizing out the data distribution $q(y_0, 0)$.
The score function is the Riemannian gradient of the log-density $\nabla_g \log q$.
The following theorem tells us that we can create a family of inference and generative SDEs that induce the same marginal distributions over $Y_s$ and $X_{T-s}$ as (\ref{eq:fixed-inference-gen}) if we have access to its score. 
\begin{mdframed}[style=MyFrame2]
\begin{restatable}[\textbf{Marginally equivalent SDEs}]{thm}{equivalent}
\label{thm:equivalent}
For $\lambda\leq1$, the marginal distributions of $X_{T-s}$ and $Y_{s}$ of the processes defined as below
\begin{align}
    \ddd Y &= \left( U_0 -   \frac{\lambda}{2} \nabla_g \log q  \right)\ds + \sqrt{1-\lambda} V \circ \ddd \hat{B}_s  & Y_0\sim q(\cdot, 0) \label{eq:equiv-lambda}\\
    \ddd X &= \left(\left(1-\frac{\lambda}{2}\right) \nabla_g \log q - U_0\right)\dt + \sqrt{1-\lambda} \circ V \ddd \hat{B}_t & X_0 \sim  q(\cdot, T) \label{eq:equiv-lambda-reverse}
\end{align}
both have the density $q(\cdot, s)$.
In particular, $\lambda=1$ gives rise to an equivalent ODE.
\end{restatable}
\end{mdframed}
This suggests if we can approximate the score function, and plug it into the  reverse process (\ref{eq:equiv-lambda-reverse}), we obtain a time-reversed process that induces approximately the same marginals. 
\begin{mdframed}[style=MyFrame2]
\begin{restatable}[\textbf{Score matching equivalency}]{thm}{score}
\label{thm:score}
For $\lambda <1$, let $\gE_\lambda^\infty$ denote the Riemannian CT-ELBO of the generative process (\ref{eq:equiv-lambda-reverse}), with $\nabla_g \log q$ replaced by an approximate score $S_\theta$, and with (\ref{eq:equiv-lambda}) being the inference SDE.
Assume $S_\theta$ is a compactly supported smooth vector. 
Then
\begin{align}
    \E_{\vy_0}[\gE^\infty_\lambda]
    &= - C_1 \int_0^T \,\E_{\vy_s}\left[ \norm{S_\theta - \nabla_g \log q}_g^2 \right]\ds + C_2
    \label{eq:continuous-elbo-ism-lambda-averaged}
\end{align}
where $C_1>0$ and $C_2$ are constants wrt $\theta$.
\end{restatable}
\end{mdframed}
The first implication of the theorem is that \emph{maximizing the Riemannian CT-ELBO of the plug-in reverse process is equivalent to minimizing the Riemannian score-matching loss}.
Second, if we set $\lambda=0$, from (\ref{eq:lambda-induced-a}) (in the appendix), we have $Va=S_\theta$, which is exactly the fixed-inference training in~\S\ref{sec:fixed-inf}. 
That is, the vector $Va$ trained using \eqref{eq:ct-elbo-time2} is actually an approximate score, allowing us to extract an equivalent ODE by substituting $Va$ for $\nabla_g \log q$ in (\ref{eq:equiv-lambda},\ref{eq:equiv-lambda-reverse}) by setting $\lambda=1$. %

\begin{figure}[!t]
    \centering
      \medskip
    \begin{subfigure}[t]{.47\linewidth} 
    \centering
    \includegraphics[width=0.98\linewidth]{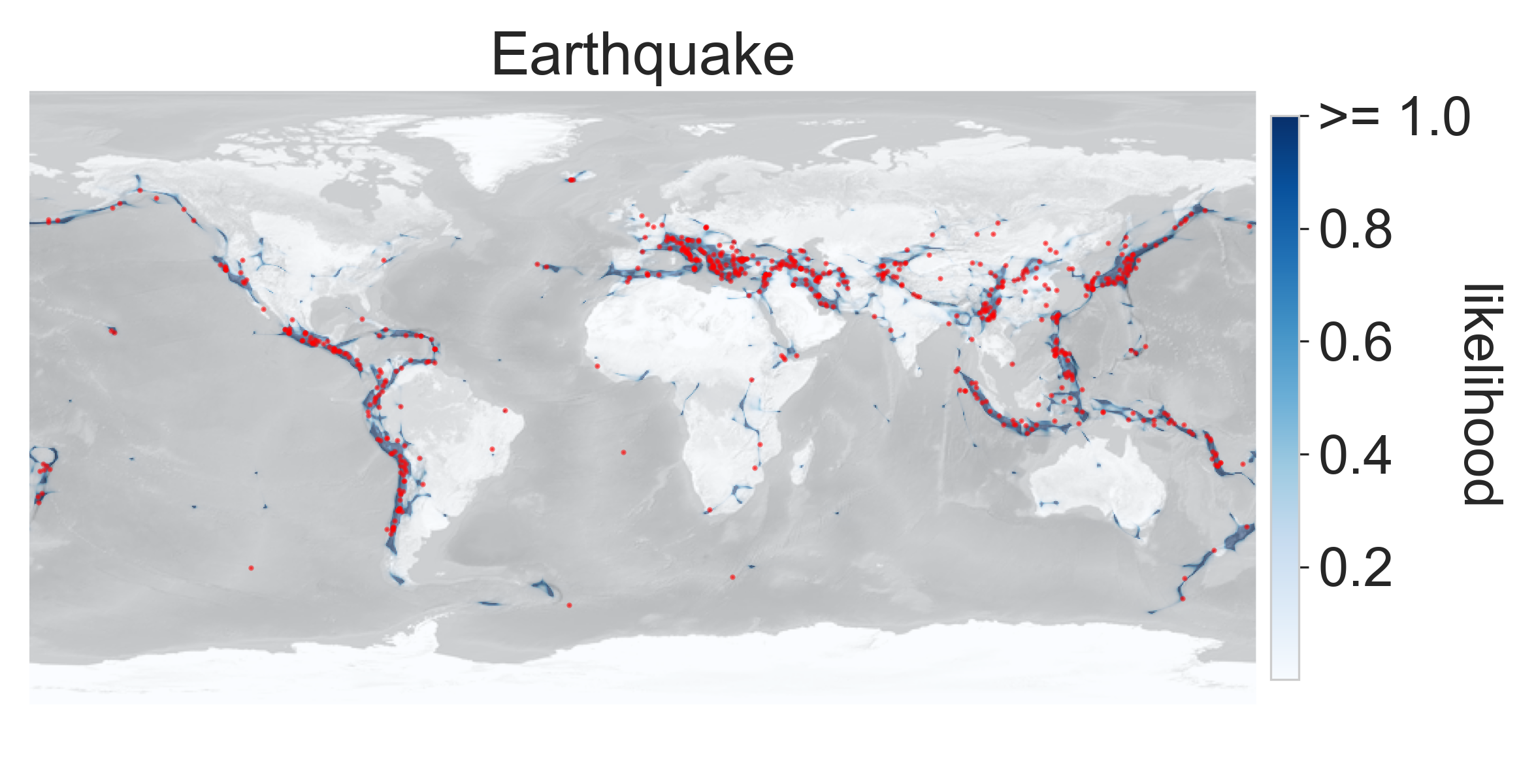}
    \end{subfigure}
    \vspace{-0.5cm}
    \begin{subfigure}[t]{.47\linewidth} 
    \centering
    \includegraphics[width=0.98\linewidth]{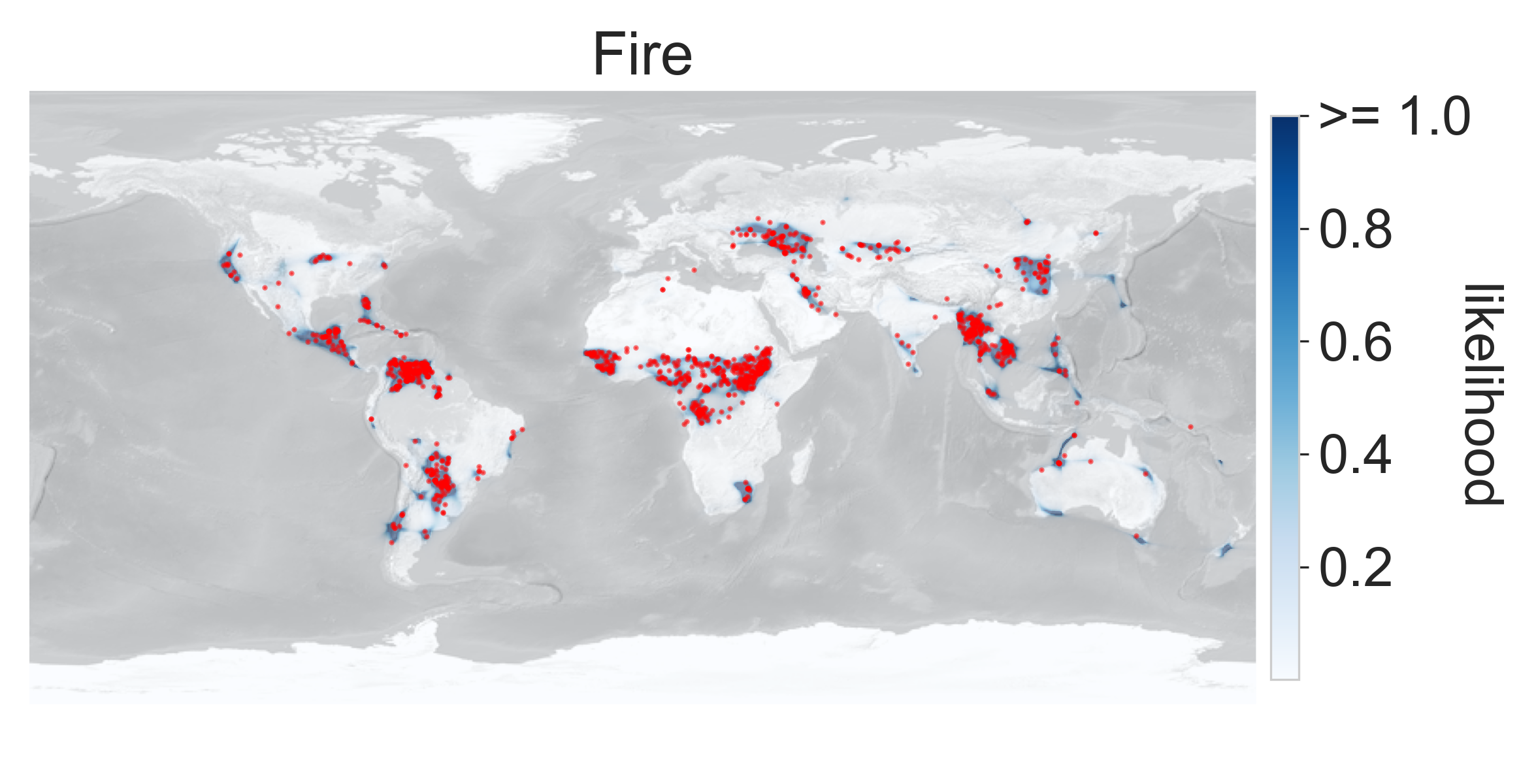}
    \end{subfigure}
    \begin{subfigure}[t]{.47\linewidth} 
    \centering
    \includegraphics[width=0.98\linewidth]{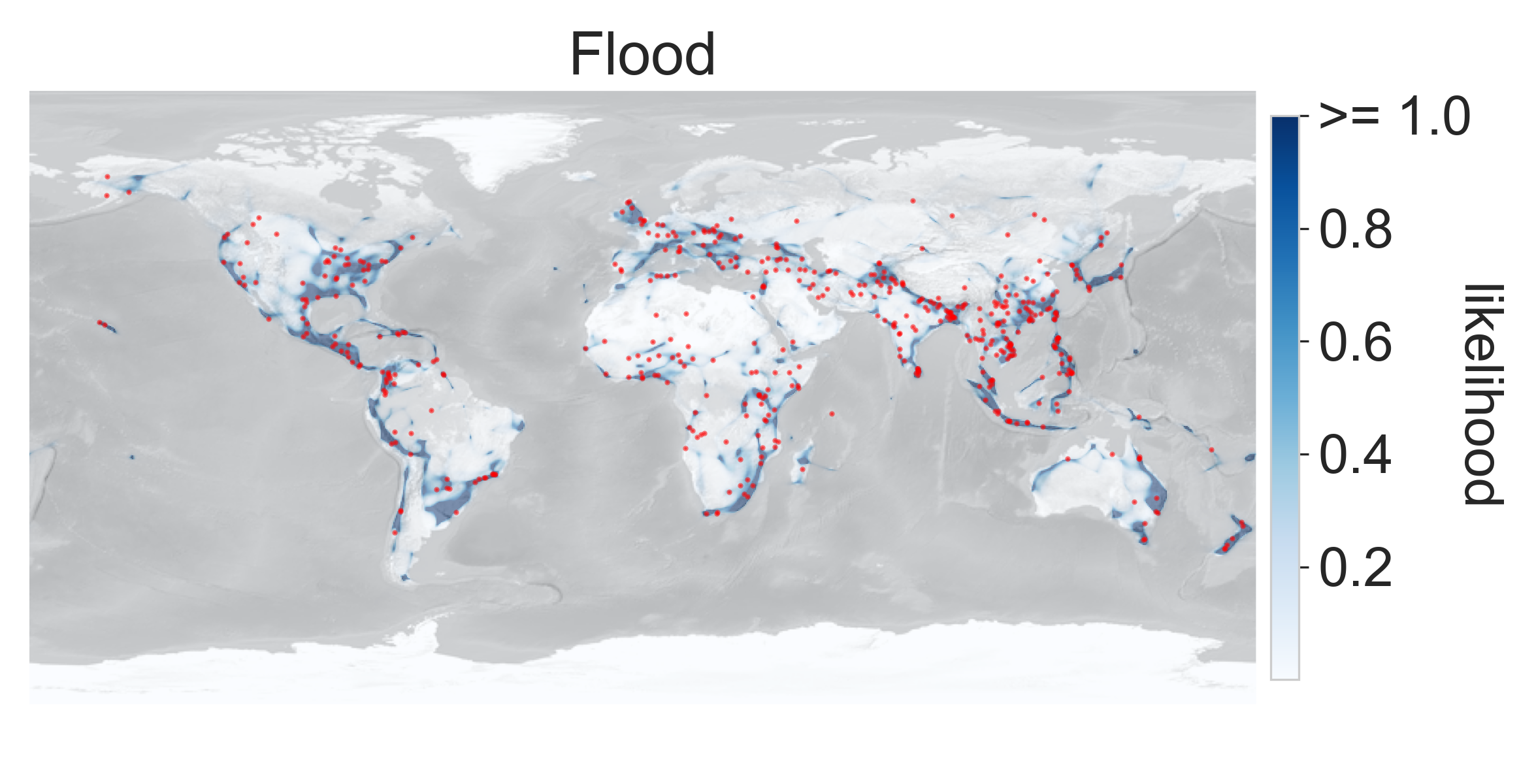}
    \end{subfigure}
    \begin{subfigure}[t]{.47\linewidth} 
    \centering
    \includegraphics[width=0.98\linewidth]{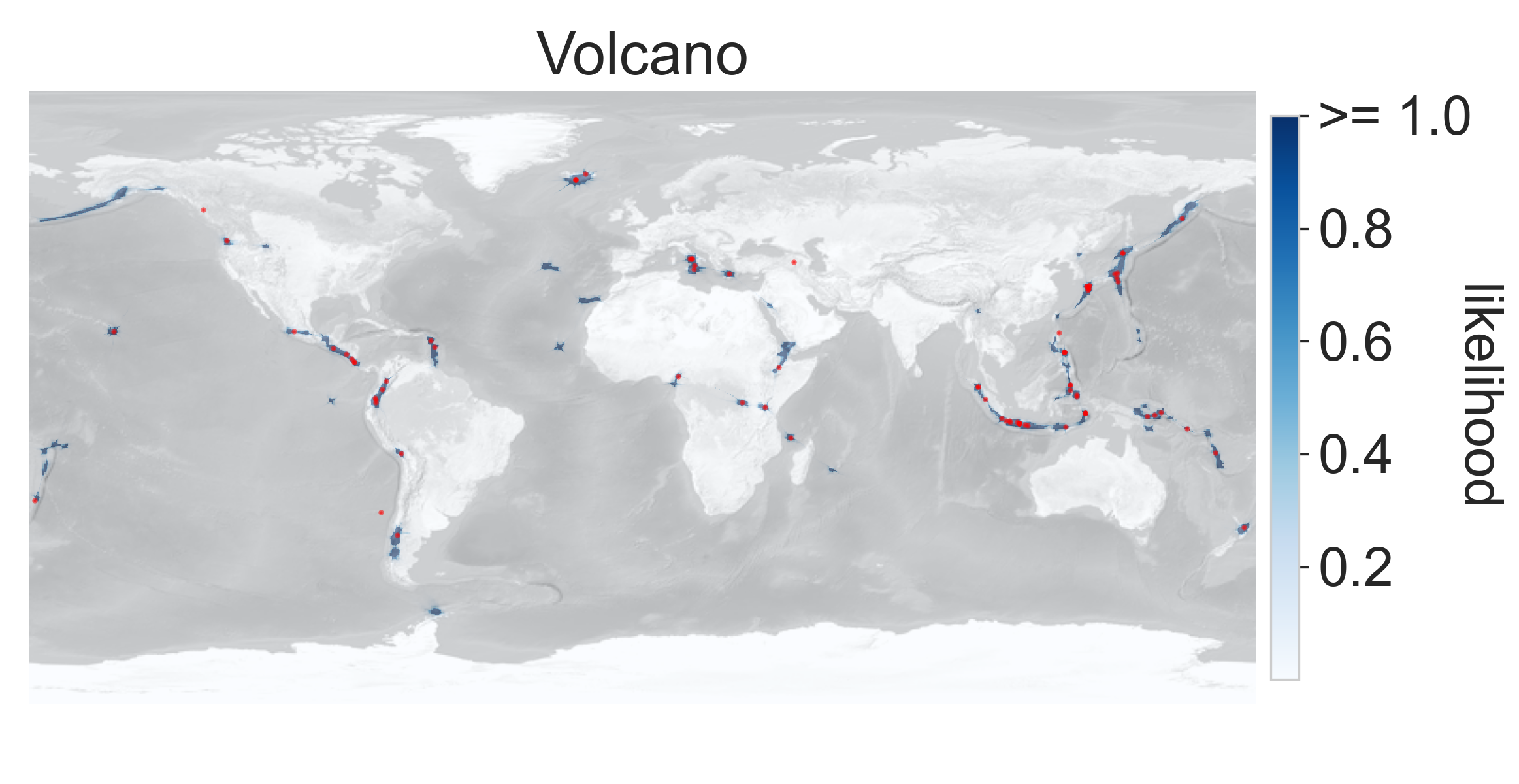}
    \end{subfigure}
    \vspace{-10pt}
    \caption{%
    \vspace{-10pt}
       \small
       Density of models trained on earth datasets. Red dots are samples from the test set.
    }%
    \vspace{-5pt}
   \label{fig:earthdensity}
\end{figure}
\section{Related work}
\vspace{-10pt}
\everypar{\looseness=-1}
\xhdr{Diffusion models} Diffusion models can be viewed from two different but ultimately complimentary perspectives. 
The first approach leverages score based generative models \citep{song2019generative,song2021score}, while the second approach treats generative modeling as inverting a fixed noise-injecting process \citep{sohl2015deep, ho2020denoising}. 
Finally, continuous-time diffusion models can also be embedded within a maximum likelihood framework \citep{huang2021variational, song2021maximum}, which represents the special case of prescribing a flat geometry---\ie Euclidean---to the generative model and is completely generalized by the theory developed in this work.

\xhdr{Riemannian Generative Models}
Generative models beyond Euclidean manifolds have recently risen to prominence with early efforts focusing on constant curvature manifolds \citep{bose2020latent,rezende2020normalizing}. 
Another line of work extends continuous-time flows \citep{chen2018neural} to more general Riemannian manifolds \citep{lou2020neural,mathieu2020riemannian,falorsi2020neural}.
To avoid explicitly solving an ODE during training, \citet{rozen2021moser} propose \emph{Moser Flow} whose objective involves computing the Riemannian divergence of a parametrized vector field.
Concurrent to our work, \citet{de2022riemannian} develop Riemannian score-based generative models for compact manifolds like the Sphere. 
While similar in endeavor, RDMs are couched within the the maximum likelihood framework. %
As a result our approach is directly amenable to variance reduction techniques via importance sampling and likelihood estimation.
Moreover, our approach is also applicable to non-compact manifolds such as hyperbolic spaces, and we demonstrate this in our experiments on a larger variety of manifolds including the orthogonal group and toroids.
\section{Experiments}
\label{sec:experiments}
We investigate the empirical caliber of RDMs on 
a range of manifolds. 
We instantiate RDMs by parametrizing $a$ in (\ref{eq:fixed-inference-gen}) using an MLP and maximize the CT-ELBO (\ref{eq:ct-elbo-time2}). 
We report our detailed training procedure---including selected hyperparameters---for all models in~\S\ref{sec:experimental-details}.

\subsection{Sphere}
For spherical manifolds, 
we model the datasets compiled by \citet{mathieu2020riemannian}, which consist of earth and climate science events on the surface of the earth such as volcanoes \citep{volcanoe_dataset}, earthquakes \citep{earthquake_dataset}, floods \citep{flood_dataset}, and fires \citep{fire_dataset}. 
In Table \ref{tab:geoscience} for each dataset we report average and standard deviation of test negative log likelihood on 5 different runs with different splits of the dataset. 
In Figure \ref{fig:earthdensity} we plot the model density in blue while the test data is depicted with red dots.

\begin{table}[t]
    \centering
    \scriptsize
    \begin{tabular}{lccccc}
     & \textbf{Volcano} & \textbf{Earthquake} & \textbf{Flood} & \textbf{Fire} \\
    \midrule
    Mixture of Kent & $-0.80_{\pm 0.47}$ & $0.33_{\pm 0.05}$ & $0.73_{\pm 0.07}$ & $-1.18_{\pm 0.06}$ \\
    Riemannian CNF \citep{mathieu2020riemannian} & $-0.97_{\pm 0.15}$ & $0.19_{\pm0.04}$ & $0.90_{\pm0.03}$ & $-0.66_{\pm0.05}$ \\
    Moser Flow \citep{rozen2021moser} & $-2.02_{\pm 0.42}$ & $-0.09_{\pm0.02}$ & $0.62_{\pm 0.04}$ & $-1.03_{\pm 0.03}$ \\
    Stereographic Score-Based & ${-4.18}_{\pm 0.30}$ & ${-0.04}_{\pm 0.11}$ & ${1.31}_{\pm 0.16}$ & $0.28_{\pm 0.20}$ \\
    Riemannian Score-Based \citep{de2022riemannian} & $-5.56_{\pm0.26}$ & $-0.21_{\pm0.03}$ & $0.52_{\pm0.02}$ & $-1.24_{\pm 0.07}$\\
    \midrule
    RDM & $\bm{-6.61}_{\pm 0.97}$ & $\bm{-0.40}_{\pm 0.05}$ & $\bm{0.43}_{\pm 0.07}$ & $\bm{-1.38}_{\pm0.05}$\\
    \midrule 
    Dataset size & 827 & 6120 & 4875 & 12809 \\
    \bottomrule
    \end{tabular}
    \caption{
    \small
    NLL scores for each method on earth datasets.
    Bold shows best results (up to statistical significance).
    Means and standard deviations are calculated over $5$ runs. Baselines taken from \citet{de2022riemannian}.
    \vspace{-0.7cm}
    }
    \label{tab:geoscience}
\end{table}

\begin{figure}[t]
    \begin{minipage}{.32\textwidth}
    \centering
    \includegraphics[width=0.98\linewidth]{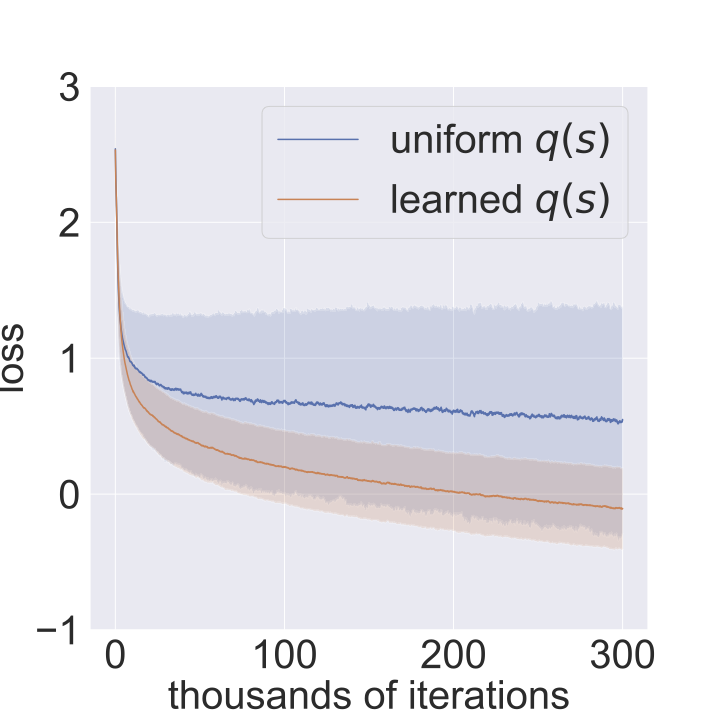}
    \caption{\small Variance reduction with importance sampling.}
    \label{fig:imp_ablation}
    \end{minipage}
    \vspace{-0.1cm}
    \hfill
    \begin{minipage}{.64\textwidth}
    \includegraphics[width=0.485\linewidth]{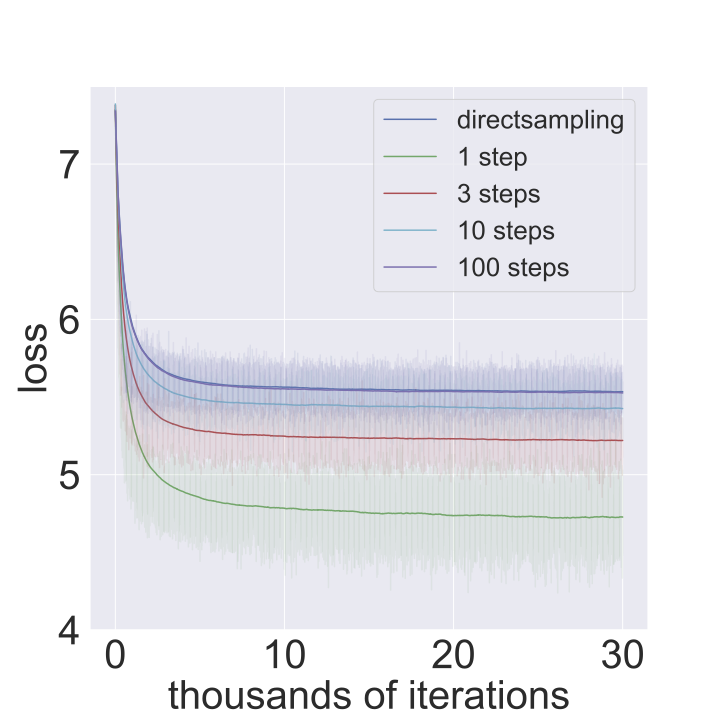}
    \hfill
    \includegraphics[width=0.485\linewidth]{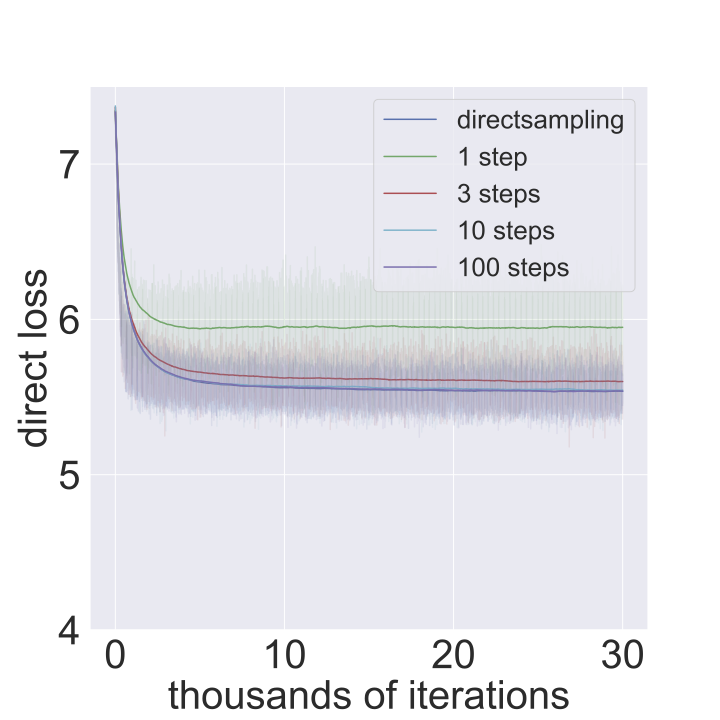}
    \caption{
    \small Direct sampling vs numerical integration of Brownian motion. Numbers in legends indicate the number of time steps. }
    \label{fig:tori_ablation}
    \end{minipage}
    \vspace{-0.1cm}
\end{figure}

\xhdr{Variance reduction}
We demonstrate the effect of applying variance reduction on optimizing the Riemannian CT-ELBO (\ref{eq:ct-elbo-time2}) using the earthquake dataset. 
As shown in Figure \ref{fig:imp_ablation}, learning an importance sampling proposal effectively lowers the variance and speeds up training.

\subsection{Tori}
\label{sec:tori}
For tori, we use the list of 500 high-resolution proteins compiled in \citet{lovell2003structure} and select 113 RNA sequences listed in \citet{murray2003rna}. 
Each macromolecule is divided into multiple monomers, and the joint structure is discarded---we model the lower dimensional density of the backbone conformation of the monomer. 
For the protein data, this corresponds to 3 torsion angles of the amino acid.
As one of the angles is normally 180°, we also discard it, and model the density over the 2D torus.
For the RNA data, the monomer is a nucleotide described by 7 torsion angles in the backbone, represented by a 7D torus. 
For protein, we divide the dataset by the type of side chain attached to the amino acid, resulting in 4 datasets, and we discard the nucleobases of the RNA.

In Table \ref{tab:tori} we report the NLL of our model. 
Our baseline is a mixture of $4,096$ power spherical distributions \citep[MoPS]{https://doi.org/10.48550/arxiv.2006.04437}. %
We observe that RDM outperforms the baseline across the board, and the difference is most noticeable for the RNA data, which has a higher dimensionality. %

\begin{figure}[!t]
    \centering
      \medskip
    \begin{minipage}[c]{0.99\linewidth}
    \includegraphics[width=0.24\linewidth, trim={0cm 0cm 0cm 1.0cm},clip]{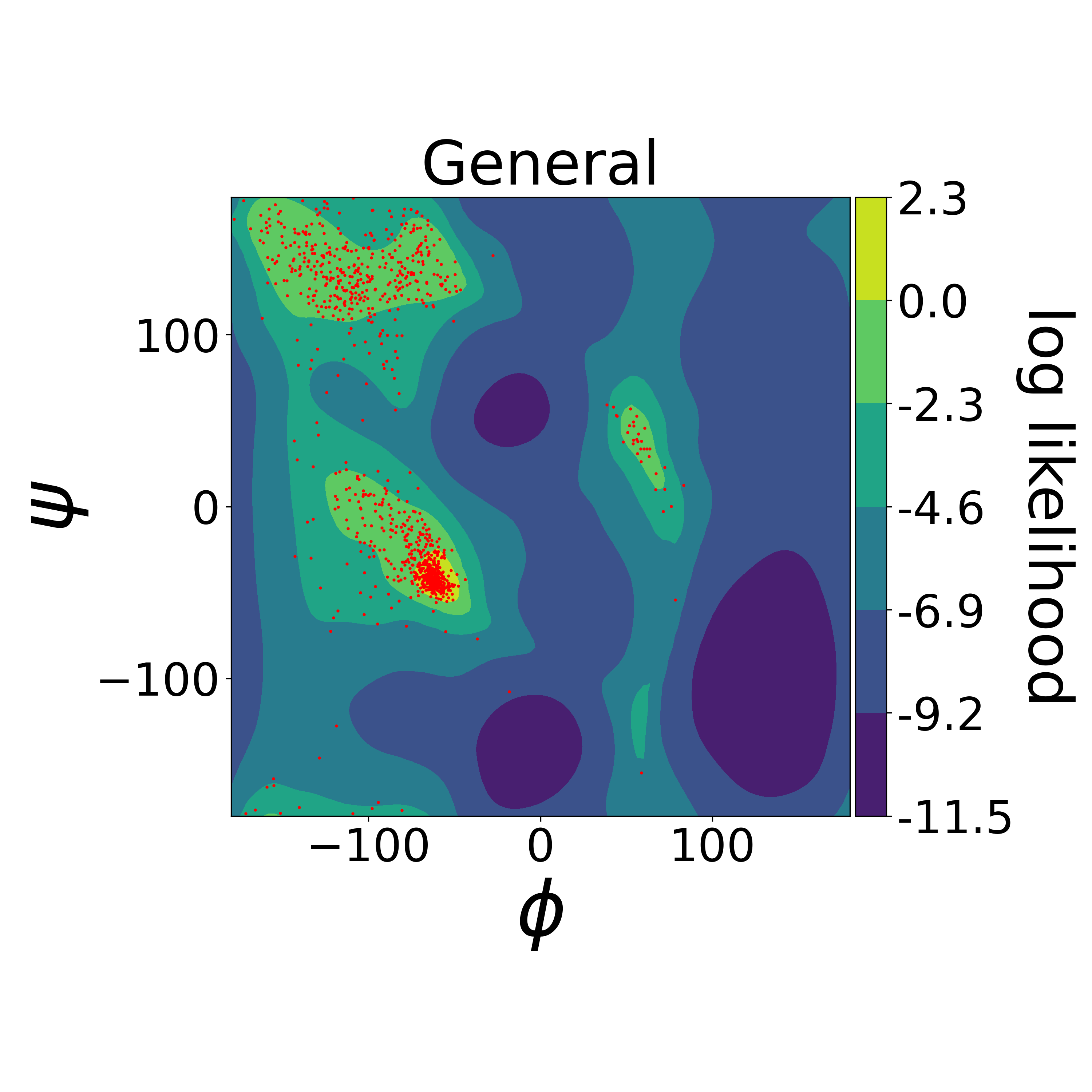}
    \includegraphics[width=0.24\linewidth, trim={0cm 0cm 0.0cm 1.0cm},clip]{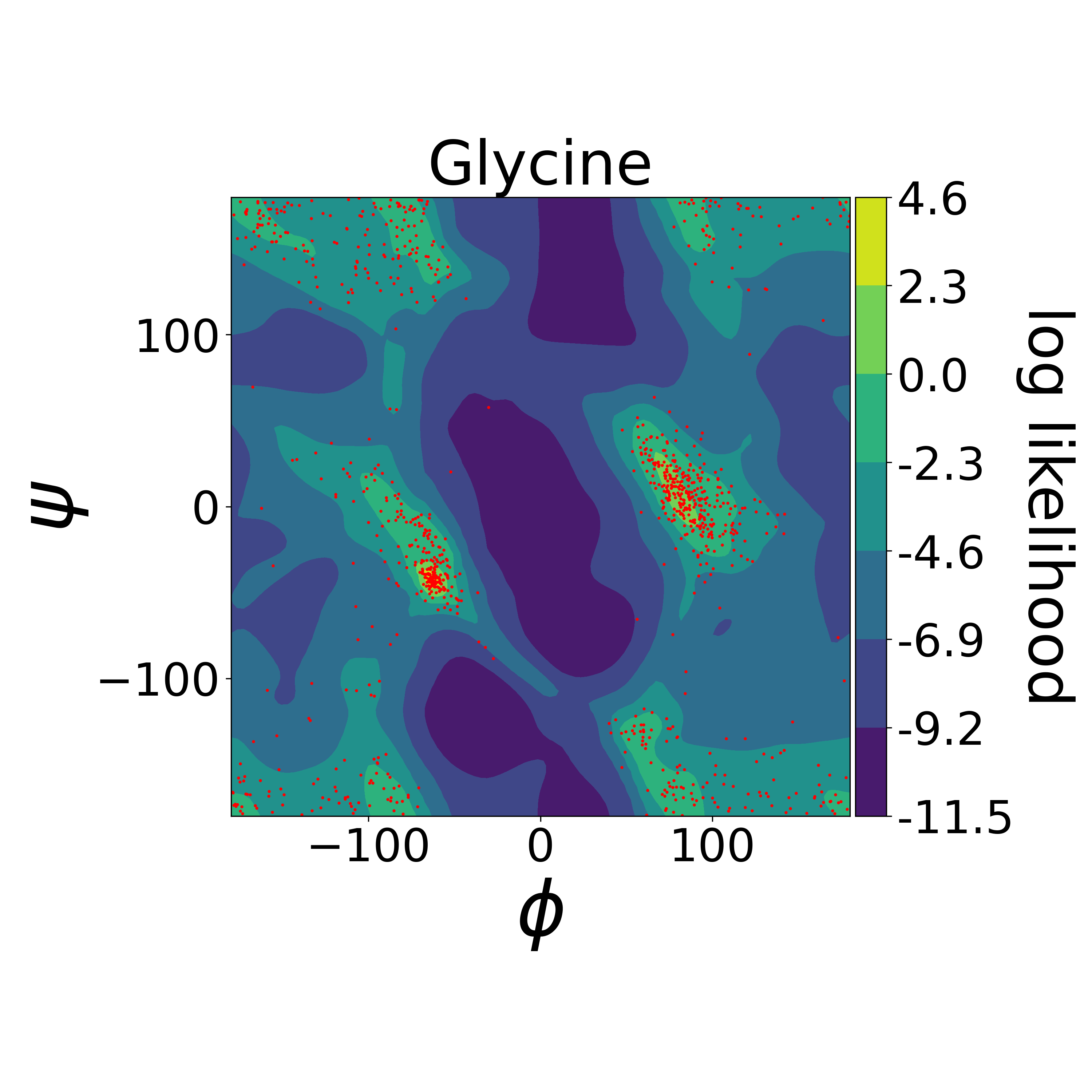}
    \includegraphics[width=0.24\linewidth, trim={0cm 0cm 0cm 1.0cm},clip]{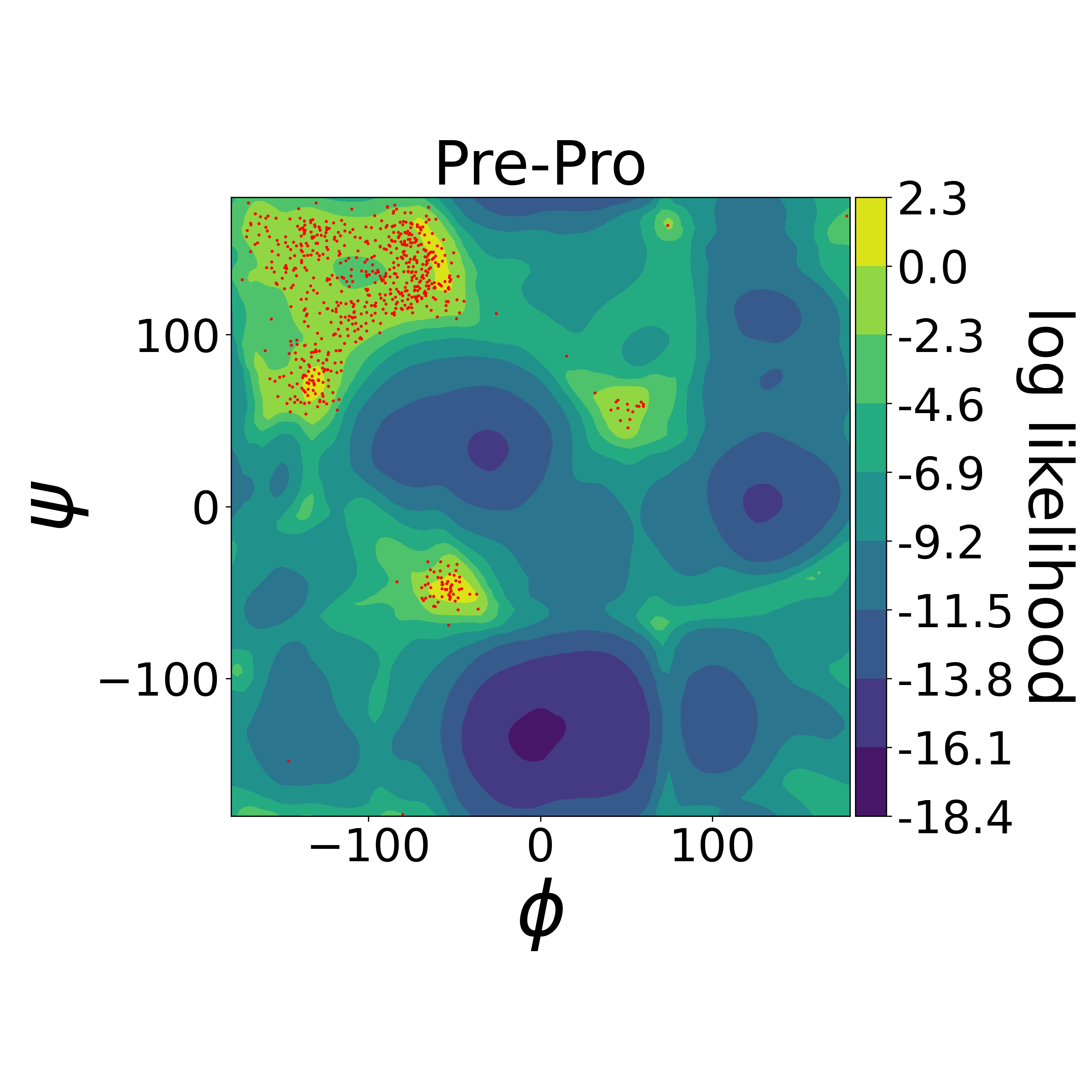}
    \includegraphics[width=0.24\linewidth, trim={0cm 0cm 0cm 1.0cm},clip]{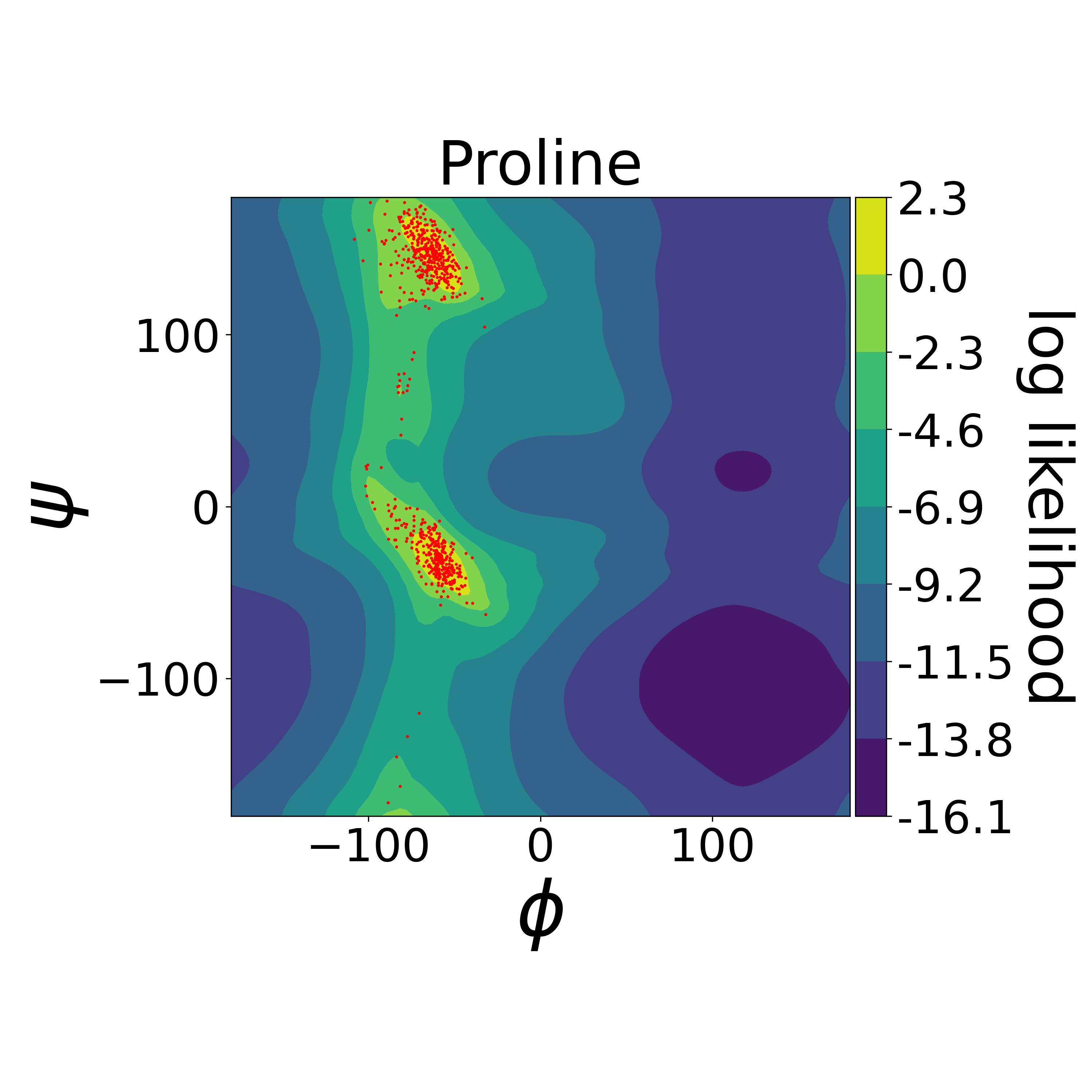}
    \end{minipage}
    \vspace{-0.3cm}
    \caption{%
      \small  Ramachandran contour plot of the model density for protein datasets. Red dots are set test samples.
     }%
   \label{fig:proteindensity}
   \vspace{-0.2cm}
\end{figure}

\cut{
Hyperbolic: non-compact manifolds.
Grassmannian / Orthogonal groups: higher dimensional problems?}

\begin{table}[t]
    \small
    \centering
    \begin{tabular}{lrrrrr}
     & \textbf{General} & \textbf{Glycine} & \textbf{Proline} & \textbf{Pre-Pro} & \textbf{RNA} \\
    \midrule
    MoPS & $1.15_{\pm 0.002}$ & $2.08_{\pm 0.009}$ & $0.27_{\pm 0.008}$ & $1.34_{\pm 0.019}$ &  $4.08_{\pm 0.368}$ \\
    RDM & $\bm{1.04}_{\pm 0.012}$ & $\bm{1.97}_{\pm 0.012}$ & $\bm{0.12}_{\pm 0.011}$ & $\bm{1.24}_{\pm0.004}$  & $\bm{-3.70}_{\pm0.592}$\\
    \midrule 
    Dataset size & 138208 & 13283 & 7634 & 6910 & 9478 \\
    \bottomrule
    \end{tabular}
    \caption{ 
    \small
    Negative test log-likelihood for each method on Tori datasets.
    Bold shows best results (up to statistical significance).
    Means and standard deviations are calculated over $5$ runs.
    \vspace{-20pt}
    }
    \label{tab:tori}
\end{table}

\xhdr{Numerical integration ablation}
We estimate the loss (\ref{eq:ct-elbo-time2}) by integrating the inference SDE on $\mathcal{M}$. 
To study the effect of integration error, we experiment with various numbers of time steps evenly spaced between $[0, s]$ on Glycine. %
Also, as we can directly sample the Brownian motion on tori without numerical integration, we use it as a reference (termed direct loss) for comparison. %
Figure \ref{fig:tori_ablation} shows while fewer time steps tend to underestimate the loss, the model trained with 100 time steps is already indistinguishable from the one trained with direct sampling.
We also find numerical integration is not a significant overhead as each experiment takes approximately the same wall-clock time with identical setups. %
This is because the inference path does not involve the neural module $a$.

\subsection{Hyperbolic Manifolds}
\begin{wrapfigure}[10]{r}{0.40\textwidth}
    \vspace{-1.2cm}
    \centering
    \includegraphics[width=0.12\textwidth]{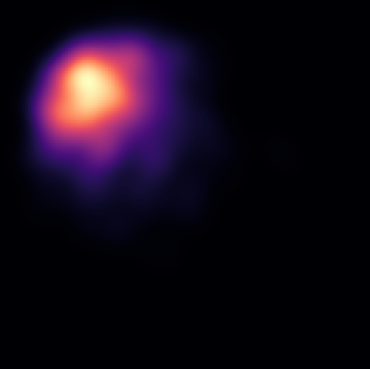} \hfill
    \includegraphics[width=0.12\textwidth]{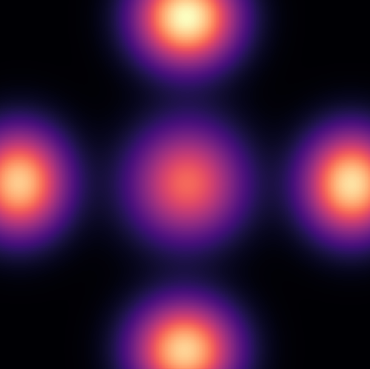} \hfill
    \includegraphics[width=0.12\textwidth]{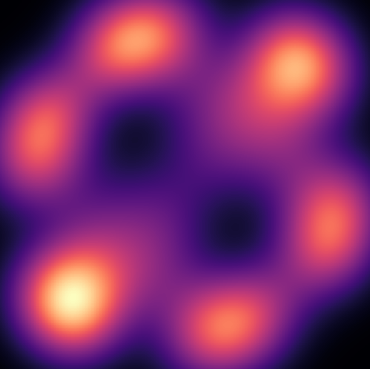} \\
    \includegraphics[width=0.12\textwidth]{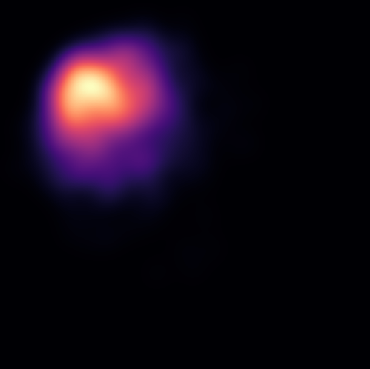} \hfill
    \includegraphics[width=0.12\textwidth]{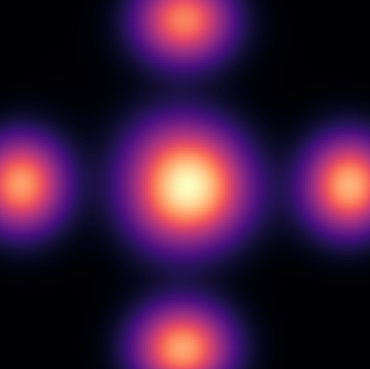} \hfill
    \includegraphics[width=0.12\textwidth]{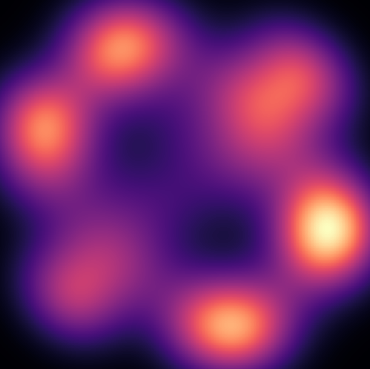}
    \caption{\small Hyperbolic Manifold. Top: data. Bottom: learned Density}
    \label{fig:hyperbolics}
\end{wrapfigure}
Hyperbolic manifolds provide an example whose closest-point projection is not cheap to obtain, and a claimed closest-point projection in recent literature is in fact not the closest \emph{Euclidean projection}~\citep{skopek2019mixed} (see~\S\ref{app:manifolds} for more details). 
To demonstrate the generality of our framework, we model the synthetic datasets in  Figure~\ref{fig:hyperbolics}, first introduced by \citet{bose2020latent,lou2020neural}.
Since hyperbolic manifolds are not compact, we need a non-zero drift to ensure the inference processs is not dissipative. 
We define the prior as the standard normal distribution on the $yz$-plane and let $U_0$ be $\frac{1}{2}\nabla_g \log p_0$, so that $Y_s$ will revert back to the origin.

\cut{
\begin{table}[h]
    \centering
    \begin{tabular}{lrrrr}
     & \textbf{1Wrapped Gaussian} & \textbf{MoG Wrapped Gaussians} & \textbf{Checkerboard}  \\
    \midrule
    NMODE &  & $1.313 \pm $ & $3.056 \pm $\\
    \midrule
    Manifold Diffusion & & & $4.941 \pm$\\
    \bottomrule
    \end{tabular}
    \caption{
    Negative log-likelihood scores for synthetic hyperbolic densities.
    }
    \label{tab:hyperbolic}
\end{table}
}

\cut{
\begin{figure}[ht!]
    \centering
      \medskip
    \begin{minipage}[c]{.99\linewidth}
    \includegraphics[width=0.24\linewidth]{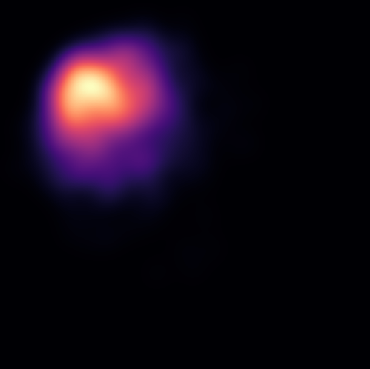}
    \includegraphics[width=0.24\linewidth]{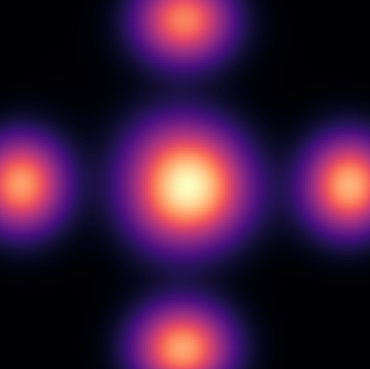}
    \includegraphics[width=0.24\linewidth]{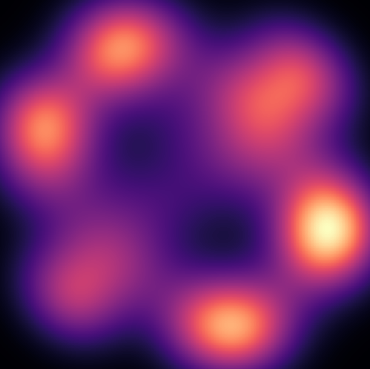}
    \includegraphics[width=0.24\linewidth]{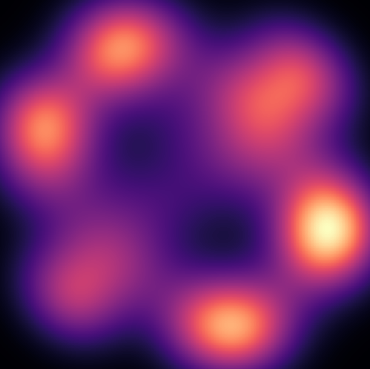} \\ 
    \includegraphics[width=0.24\linewidth]{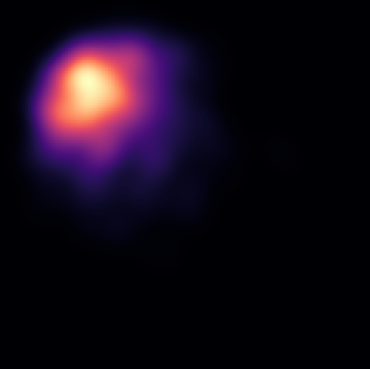}
    \includegraphics[width=0.24\linewidth]{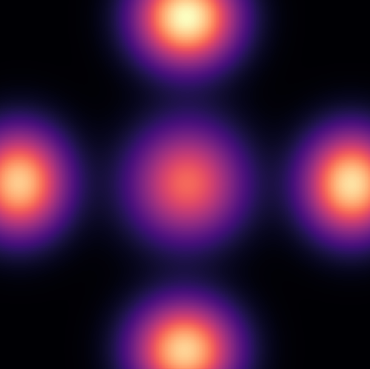}
    \includegraphics[width=0.24\linewidth]{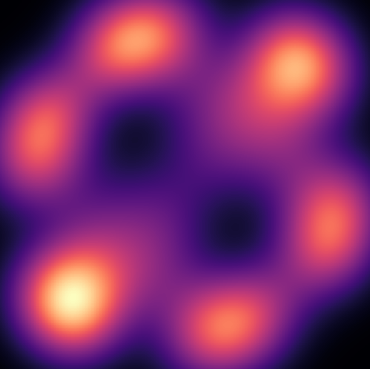}
    \includegraphics[width=0.24\linewidth]{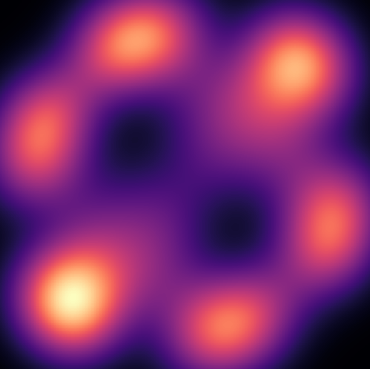}
    \end{minipage}
    \caption{%
       TODO
     }%
   \label{fig:hyperbolic_densities}
\end{figure}
}

\subsection{Special Orthogonal Group}
\begin{wrapfigure}[8]{r}{0.4\textwidth}
    \vspace{-1.4cm}
    \centering
    \includegraphics[width=0.18\textwidth, trim={10cm 5cm 15.0cm 5.0cm},clip]{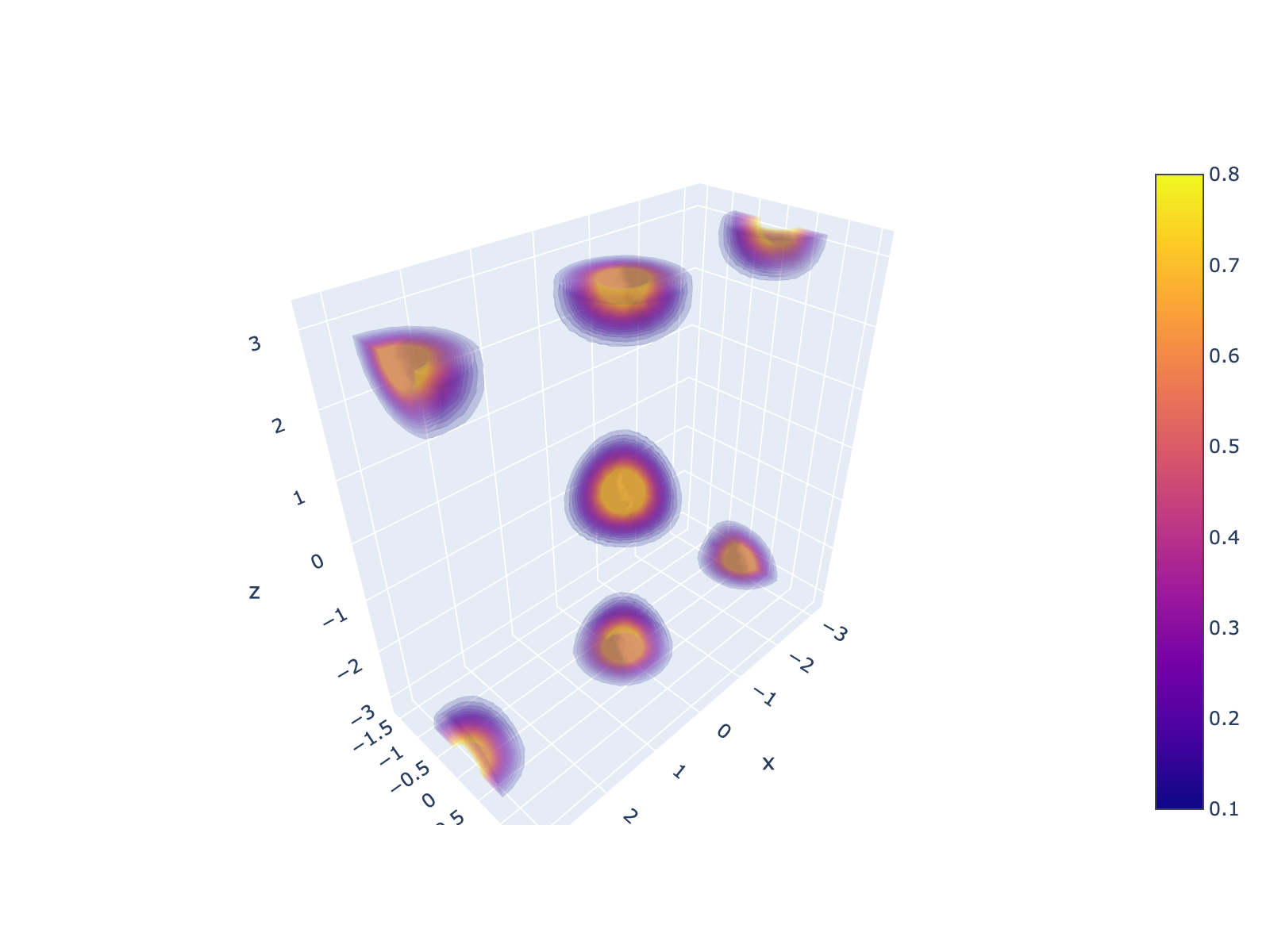}
    \includegraphics[width=0.18\textwidth, trim={10cm 5cm 15.0cm 5.0cm},clip]{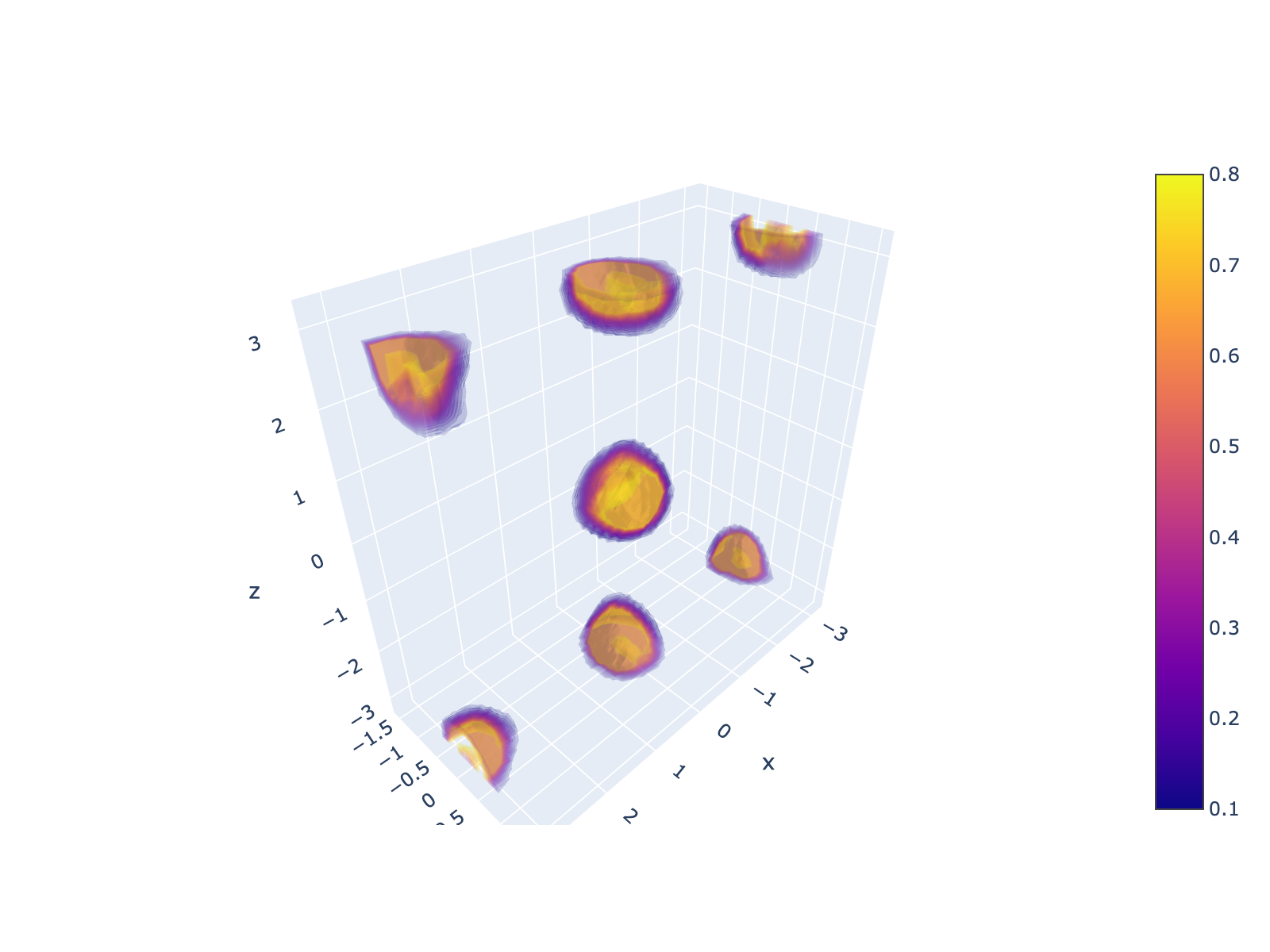}
    \caption{\small SO(3). Left: synthetic multimodal density. Right: learned density.}
    \label{fig:so3}
\end{wrapfigure}
Another example whose closest-point projection is expensive to compute is the orthogonal group, as it requires performing the singular value decomposition. 
To evaluate our framework on this matrix group, we generate data using the synthetic multimodal density defined on $SO(3)$ from \citet{brofos2021manifold}.
We view it as a submanifold embedded in $\R^{3\times3}$, therefore $d=3$ and $m=9$.  
We use the projected Hutchinson to estimate the Riemannian divergence.
Since the data are 3D rotational matrices, we can visualize them using the \emph{Euler angles}. 
We plotted the data density and the learned model density in Figure~\ref{fig:so3}, where each coordinate represents the rotation around that particular axis.

\section{Conclusion}
\everypar{\looseness=-1}
In this paper, we introduce RDMs that extend continuous-time diffusion models to arbitrary Riemannian manifolds---including challenging non-compact manifolds like hyperbolic spaces. 
We provide a variational framework to train RDMs by optimizing a novel objective, the Riemannian Continuous-Time ELBO. 
To enable efficient and stable training we provide several key tools such as a fixed-inference paramterization of the SDE in the ambient space, new methodological techniques to compute the Riemannian divergence, 
as well as an importance sampling procedure with respect to the time integral to reduce the variance of the loss. 
On a theoretical front, we also show deep connections between our proposed variational framework and Riemannian score matching through the construction of marginally equivalent SDEs. 
Finally, we complement our theory by constructing RDMs that achieve state-of-the-art performance on density estimation on geoscience datasets,  protein/RNA data on toroidal, and synthetic data on hyperbolic and orthogonal-group manifolds.

\clearpage
\bibliography{ref}
\bibliographystyle{icml_2021}

\section*{Checklist}

The checklist follows the references.  Please
read the checklist guidelines carefully for information on how to answer these
questions.  For each question, change the default \answerTODO{} to \answerYes{},
\answerNo{}, or \answerNA{}.  You are strongly encouraged to include a {\bf
justification to your answer}, either by referencing the appropriate section of
your paper or providing a brief inline description.  For example:
\begin{itemize}
  \item Did you include the license to the code and datasets? \answerYes{See Section~\ref{gen_inst}.}
  \item Did you include the license to the code and datasets? \answerNo{The code and the data are proprietary.}
  \item Did you include the license to the code and datasets? \answerNA{}
\end{itemize}
Please do not modify the questions and only use the provided macros for your
answers.  Note that the Checklist section does not count towards the page
limit.  In your paper, please delete this instructions block and only keep the
Checklist section heading above along with the questions/answers below.

\begin{enumerate}

\item For all authors...
\begin{enumerate}
  \item Do the main claims made in the abstract and introduction accurately reflect the paper's contributions and scope?
    \answerYes
  \item Did you describe the limitations of your work?
    \answerYes{There are a few algorithmic choices we made (limited by the framework and the fact we have a dynamical model) such as numerical integration, and divergence computation / estimation. 
    Our ablation studies indicate that it does not pose a big problem at least for the problems we are looking at. 
    However, that does not mean it can be applied to other settings (like higher dimensional problems) out of the box though.}
  \item Did you discuss any potential negative societal impacts of your work?
    \answerNA
  \item Have you read the ethics review guidelines and ensured that your paper conforms to them?
    \answerYes
\end{enumerate}

\item If you are including theoretical results...
\begin{enumerate}
  \item Did you state the full set of assumptions of all theoretical results?
    \answerYes
        \item Did you include complete proofs of all theoretical results?
    \answerYes
\end{enumerate}

\item If you ran experiments...
\begin{enumerate}
  \item Did you include the code, data, and instructions needed to reproduce the main experimental results (either in the supplemental material or as a URL)?
    \answerYes
  \item Did you specify all the training details (e.g., data splits, hyperparameters, how they were chosen)?
    \answerYes
        \item Did you report error bars (e.g., with respect to the random seed after running experiments multiple times)?
    \answerYes
        \item Did you include the total amount of compute and the type of resources used (e.g., type of GPUs, internal cluster, or cloud provider)?
    \answerYes
\end{enumerate}

\item If you are using existing assets (e.g., code, data, models) or curating/releasing new assets...
\begin{enumerate}
  \item If your work uses existing assets, did you cite the creators?
    \answerYes
  \item Did you mention the license of the assets?
    \answerNA
  \item Did you include any new assets either in the supplemental material or as a URL?
    \answerNA
  \item Did you discuss whether and how consent was obtained from people whose data you're using/curating?
    \answerNo
  \item Did you discuss whether the data you are using/curating contains personally identifiable information or offensive content?
    \answerNo
\end{enumerate}

\item If you used crowdsourcing or conducted research with human subjects...
\begin{enumerate}
  \item Did you include the full text of instructions given to participants and screenshots, if applicable?
    \answerNA
  \item Did you describe any potential participant risks, with links to Institutional Review Board (IRB) approvals, if applicable?
    \answerNA
  \item Did you include the estimated hourly wage paid to participants and the total amount spent on participant compensation?
    \answerNA
\end{enumerate}

\end{enumerate}

\newpage
\appendix
\section{Riemannian manifolds}
\label{app:riem}
\xhdr{Notation Convention} There are a number of different notations used in differential geometry and all have their place. 
The most abstract level is with tensors (of which forms and vectors are special cases) and it is the best for establishing general properties. We used an index-free notation in the main paper. 
A coordinate-based, but still intrinsic, description that uses local charts which describe explicit coordinate systems for \emph{patches} of the manifold: for the most part we use intrinsic coordinates in the main paper.  
For computational purposes it is convenient to view manifolds as hypersurfaces embedded in $\R^m$ even though this obscures the geometric meaning; these are called extrinsic coordinates which we use for actual implementations. %

We use capital letters to denote vectors, and tilded letters to denote vectors and variables defined on the local patch.

\subsection{Smooth manifolds and tangent vectors}

We recall some preliminaries of smooth manifolds.  See~\citet{lee2013smooth} for
a more detailed and comprehensive account.

A smooth $d$-manifold is a topological space $\gM$ (assumed to be paracompact, Hausdorff and second countable) and a family of pairs $\{(U_i,\varphi_i)\}$, where the $U_i$ are open sets that together cover all of $\gM$ and each $\varphi_i$ is a homeomorphism from $U_i$ to an open set in $\R^d$; these pairs are called \emph{charts}. 
They are required to satisfy a compatibility condition: if $U_i$ and $U_j$ have non-empty intersection, say $V$, then $\varphi_i\circ\varphi_j^{-1}|_V$ has to be an infinitely differentiable map from $\varphi_j(V)\subset \R^d$ to $\varphi_i(V)\subset \R^d$.  
The use of charts allows one to talk about differentiability of functions or vectors fields, by moving to $\R^d$ as needed.  
A \emph{smooth function} $f$ on $\gM$ has type $\gM\to\R$ and is such that for any chart $(U,\varphi)$ the map $f\circ\varphi^{-1}:\R^d\to\R$ is smooth \footnote{Strictly speaking this map has to be restricted to $\varphi(U)$ but we will assume that the appropriate restrictions are always intended rather than cluttering up the notation with restrictions all the time.}.  
The set of smooth functions on $\gM$ is denoted $C^\infty(\gM)$.

Let $\gM$ be a smooth manifold, and fix a point $x$ in $\gM$.
A \textbf{derivation} at $x$ is a linear operator $D:C^\infty(\gM)\rightarrow
\R$ satisfying the product rule 
\begin{align}
    D(fg) = f(x) D(g) + g(x) D(f)
\end{align}
for all $f,g\in C^\infty(\gM)$.
The set of all derivations at $x$ is a $d$-dimensional real vector space called the \textbf{tangent space} $\gT_x\gM$, and the elements of $\gT_x\gM$ are called the \textbf{tangent vectors} (or tangents) at $x$.  
For the Euclidean space $\gM=\R^d$, we have that 
$\gT_x\R^d = \textnormal{span}\{\frac{\partial}{\partial x_1},\cdots, \frac{\partial}{\partial x_d}\}$. 
We now see how to use the Euclidean derivations to induce the tangent space of arbitrary Riemannian manifolds. 

Let $\gN$ be another smooth manifold. 
For any tangent $V\in \gT_x\gM$ and smooth map $\varphi:\gM\rightarrow\gN$, the
\textbf{differential} $d \varphi_x : \gT_x\gM\rightarrow \gT_{\varphi(x)}\gN$ is defined as the pushforward of $V$ acting on $f\in C^\infty(N)$: 
\begin{align}
    d \varphi_x(v) (f) = V(f\circ \varphi).
\end{align}

Note that, if $\varphi$ is a diffeomorphism, $d \varphi_x$ is an isomorphism between $\gT_x\gM$ and $\gT_{\varphi(x)}\gN$, and the inverse map satisfies $(d \varphi_x)^{-1} = d(\varphi^{-1})_{\varphi(x)}$.  
Furthermore, differentials follow the chain rule, \ie the differential of a composite is the composite of the differentials. 

Let $\tx=(\tx_1,\cdots,\tx_d) = \varphi(x)$ be a local coordinate. 
Since $d\varphi_x:\gT_x\gM\rightarrow \gT_{\varphi(x)}\R^d$ is an isomorphism, we can characterize $\gT_x\gM$ via inversion. 
We define the basis vector $\tilde{E}_i$ of $\gT_x\gM$ by
\begin{align}
    \tilde{E}_i
    = (d\varphi_x)^{-1} \left(\frac{\partial}{\partial \tx_i}\right)
    = (d\varphi^{-1})_{\varphi(x)} \left(\frac{\partial}{\partial \tx_i}\right),
\end{align}
which means
\begin{align}
    \tilde{E}_i(f) = \frac{\partial}{\partial \tx_i}f(\varphi^{-1}(\tx)).
\end{align}
The tangent space $\gT_x\gM$ of $\gM$ at $x$ is spanned by $\left\{\tilde{E}_1, \cdots, \tilde{E}_d\right\}$. 
This means any tangent vector $V$ can be represented by $\sum_{i=1}^d \tilde{v}_i \tilde{E}_i$ for some coordinate-dependent coefficients $\tilde{v}_i$.

A manifold $\gM$ is said to be \emph{embedded} in $\R^m$ if there is an inclusion map $\iota:\gM\to\R^m$ such that $\gM$ is homeomorphic to $\iota(\gM)$ and the differential at every point is injective.  
Every smooth manifold can be embedded in some $\R^m$ with $m > d$ for some suitably chosen $m$. 

When $\gM$ is embedded in $\R^m$,
we can view $\gT_x\gM$ as a linear subspace of $\gT_x\R^m$; note that this map has
trivial kernel.  
Let $\iota:\gM\rightarrow\R^m$ denote the inclusion map, \ie $\iota(x)=x\in\R^m$
for $x\in\gM$.  
Then
\begin{align}
    \tilde{E}_i
    &= (d\varphi^{-1})_{\varphi(x)} \left(\frac{\partial}{\partial \tx_i}\right)
    = (d \iota^{-1})_{\iota(x)}(d\iota\circ\varphi^{-1})_{\varphi(x)} \left(\frac{\partial}{\partial \tx_i}\right) = \sum_{j=1}^m \frac{\partial \varphi^{-1}_j}{\partial \tx_i} \frac{\partial}{\partial x_j}.
\end{align}
This means we can rewrite a tangent vector using the ambient space's basis
\begin{align}
    \sum_{i=1}^d \tilde{v}_i \tilde{E}_i  
    = \sum_{i=1}^d \sum_{j=1}^m  \tilde{v}_i \frac{\partial \varphi^{-1}_{j}}{\partial \tx_i} \frac{\partial}{\partial x_{j}} 
    = \sum_{j=1}^m  \bar{v}_{j}\frac{\partial}{\partial x_{j}}
\end{align}
where $\bar{v}_{j} = \sum_{i=1}^d \tilde{v}_i \frac{\partial \varphi^{-1}_{j}}{\partial \tx_i}$ is the coefficient corresponding to the $j$'th ambient space coordinate. What exactly is $\varphi_{j}^{-1}$?  Note that $\iota\circ(\varphi^{-1})$ is a map from $\R^d$ to $\R^m$ and it takes $\varphi(x)$ to $\iota(x)$.  It is this that we are writing as $\varphi_{j}^{-1}$.  

In matrix-vector form, we can write $\bar{v}=\frac{\ddd \varphi^{-1}}{\ddd \tx} \tilde{v}$, where $\bar{v}$ is a vector that represents the $m$-dimensional coefficients in the ambient space. 
This also means $\bar{v}$ lies in the linear subspace spanned by the column vectors of the Jacobian $\frac{\partial\varphi^{-1}}{\partial \tx_i}$. 
This linear subspace is isomorphic to $\gT_x\gM$, which itself is a subspace of $\gT_x \R^m$.
We refer to this linear subspace as the \textbf{tangential linear subspace}
Intuitively, this means a particle traveling at speed $\bar{v}$ and position $x$ can only move tangentially on the surface.
Therefore it is restricted to move on the manifold. 

A vector field $V$ is a continuous map that assigns a tangent vector to each point on the manifold; that is $V(x) \in \gT_x\gM$.
We abuse the notation a bit and use capital letters to denote both vector fields and vectors. It should be clear in the context whether it it meant to be a function of points on the manifold or not.
Such a vector field can also map a
smooth function to a function, via the assignment $x\in\gM\mapsto V(x)(f)\in\R$.
If it maps smooth functions to smooth functions we say that the vector field is smooth.  The space of smooth vector fields on
$\gM$ is denoted by by $\mathfrak{X}(\gM)$.

\subsection{Riemannian metric}

A Riemannian manifold $(\gM, g)$ is a $d$-dimensional smooth manifold $\gM$
equipped with an inner product $g_x: \gT_x\gM \times \gT_x\gM\rightarrow\R$ on the
tangent space of each $x\in\gM$ \citep{lee2018riemannian}.  $g_x$ is called the
metric tensor at $x$.

A \textbf{metric tensor field} is an assignment of a metric tensor to each point
$x$ of $\gM$; we denote it by $g$.  The metric tensor field $g$ is said to be
\textbf{smooth} if for any smooth vector fields $u$ and $v$, $g(U,V)(x) = g_x(U(x), V(x))$
is a smooth function of $x$.  When it is clear from the context, we suppress the
subscript for simplicity.  Since $g$ is an inner product, we also write
$g(u, v) = \langle u,v\rangle_g$.

The Euclidean metric $\bar{g}$ for $\R^m$ is defined as the Euclidean inner product, characterized by the delta function 
\begin{align}
\left\langle\frac{\partial}{\partial x_i}, \frac{\partial}{\partial x_j}\right\rangle=\delta_{ij},
\end{align}
which is equal to $1$ if $i=j$; otherwise it is equal to $0$.
This means for any $U,V\in T_x\gM$,
\begin{align}
    \langle U,V\rangle_{\bar{g}}
    = \left\langle \sum_{i=1}^m \bar{u}_i \frac{\partial}{\partial x_i} , \sum_{j=1}^m
    \bar{v}_j \frac{\partial}{\partial x_j}\right\rangle_{\bar{g}}  
    = \sum_{i=1}^m \bar{u}_i \bar{v}_i = \bar{u}^\top \bar{v}.
    \label{eq:euc-inner-product}
\end{align}

Generally, given a set of basis vectors, such as $\tilde{E}_i$, the metric tensor can be
represented in a matrix form, via 
\begin{align}
    g_{ij} := \langle \tilde{E}_i, \tilde{E}_j \rangle_g
\end{align}
This allows us to write the metric using the patch coordinates %
\begin{align}
    \langle U, V \rangle_g
    = \sum_{i,j} \tilde{u}_i \tilde{v}_j \langle \tilde{E}_i, \tilde{E}_j \rangle_g
    = \sum_{i,j} \tilde{u}_i \tilde{v}_j g_{ij}
    = \tilde{u}^\top G \tilde{v}
    \label{eq:riem-inner-product}
\end{align}
where $G$ is a matrix whose $i$'th row and $j$'th column corresponds to $g_{ij}$. %

Using the components of the metric tensor, we can define the \textbf{dual basis}
$\tilde{E}^i = \sum_j g^{ij} \tilde{E}_j$, where $g^{ij}$ stands for the
$(i,j)$'th entry of the inverse matrix $G^{-1}$.  
$(\tilde{E}^1,\cdots,\tilde{E}^d)$ is called the dual basis for
$(\tilde{E}_1,\cdots,\tilde{E}_d)$ since they form a bi-orthogonal system,
meaning 
\begin{align}
    \langle \tilde{E}^i, \tilde{E}_j \rangle_g
    = \left\langle \sum_k g^{ik} \tilde{E}_k, \tilde{E}_j \right\rangle_g
    = \sum_k g^{ik} \langle \tilde{E}_k, \tilde{E}_j \rangle_g
    = \sum_k g^{ik} g_{kj} 
    = (G^{-1} G)_{ij} = \delta_{ij}.
\end{align}

If $\gM$ is a submanifold, \eg if it is embedded in an ambient space, it
automatically inherits the ambient manifold's metric.  
Suppose $\gM\subset \R^m$, where $m>d$ is the dimensionality of the ambient space. 
Then $g= \iota^* \bar{g}$ is a metric induced by the inclusion map, defined by
$$g_x(u, v) = \bar{g}(d\iota_x(u), d\iota_x (v)).$$
Unwinding the definitions, we have
\begin{align}
    g_{ij}
    = \left\langle d\iota_x(\tilde{E}_i), d\iota_x(\tilde{E}_j)\right\rangle_{\bar{g}} 
    = \left\langle \sum_{k=1}^m \frac{\partial \varphi^{-1}_k}{\partial \tx_i} \frac{\partial}{\partial x_k}, \sum_{k'=1}^m \frac{\partial \varphi^{-1}_{k'}}{\partial \tx_j} \frac{\partial}{\partial x_{k'}} \right\rangle_{\bar{g}} 
    = \sum_{k=1}^m \frac{\partial \varphi^{-1}_k}{\partial \tx_i} \frac{\partial \varphi^{-1}_k}{\partial \tx_j}.
\end{align}
That is, if $\psi=\varphi^{-1}$ is the inverse map of $\varphi$, we can write $G={\frac{\ddd \psi}{\ddd \tx}}^\top \frac{\ddd \psi}{\ddd \tx}$, which can be equivalently deduced from equating (\ref{eq:euc-inner-product}) and (\ref{eq:riem-inner-product}).

An important use of the metric is to define a measure over measurable subsets of the manifold. 
Let $(U, \varphi)$ be a chart and consider all functions smooth functions $f$ supported in $U$. 
Then 
$$f\mapsto \int_{\varphi(U)} (f \sqrt{|\det G|})\circ \varphi^{-1} \,\ddd \tx$$ 
is a positive linear functional. 
Since $\gM$ is Hausdorff and locally compact, by the \emph{Riesz representation theorem} \citep[Theorem 2.14]{rudin1987real}, there exists a unique Borel measure $\mu_g$ (over $U$) such that $\int_U f \,\ddd\mu_g$ is equal to the evaluation of the functional above. 
We can then apply a partition of unity \citep[Theorem 2.23]{lee2013smooth} to extend this construction of $\mu_g$ to be defined over the entire $\gM$, which says that for any open cover $\{U_i\}$ of $\gM$, there exists a set of continuous functions $\Phi_i$ satisfying the following properties:
\begin{enumerate}
    \item $0\leq \Phi_i(x)$ for all $x\in\gM$.
    \item $\supp \Phi_i \subseteq U_i$.
    \item $\sum_{i} \Phi_i(x)=1$ for all $x\in\gM$.
    \item Any $x\in\gM$ has a neighborhood that intersects with only finitely many $\supp \Phi_i$.
\end{enumerate}

By means of the partition, we can consider the following positive linear functional instead:
\begin{align}
    f\in C_c(\gM) \mapsto \sum_{i} \int_{\varphi(U_i)} (\Psi_i f \sqrt{|\det G|})\circ \varphi^{-1} \,\ddd \tx,
\end{align}
which is always well-defined since $f$ is compactly supported in $\gM$ (only finitely many summands are non-zero). 
$\sqrt{|\det G|}$ is called the \textbf{volume density}.  
We write $|G|=|\det G|$ for short. 
A probability density $p$ over $\gM$ can be thought of as a non-negative integrable function satisfying $\int_\gM p\,\ddd\mu_g = 1$.

\subsection{Riemannian gradient and divergence} %
\label{label:riem-grad}

\paragraph{Riemannian gradient}
Another crucial structure closely related to the metric is the
\textbf{Riemannian gradient}.
The definition of Riemannian gradient $\nabla_g: f\in C^\infty(\gM) \mapsto
\nabla_g f \in \mathfrak{X}(\gM)$ is motivated by the directional derivative in
Euclidean space, satisfying  
\begin{align}
    \left\langle \nabla_g f, V \right\rangle_g = V(f)
    \label{eq:riem-directional}
\end{align}
for any $V\in\mathfrak{X}(\gM)$.

To obtain an explicit formula for the Riemannian gradient, we expand both sides
of (\ref{eq:riem-directional}): 
\begin{align}
    \left\langle \nabla_g f, V \right\rangle_g
    = \sum_{i,j=1}^d \tilde{u}_i \tilde{v}_j g_{ij}
\end{align}
where we let $\tilde{u}_i$ and $\tilde{v}_j$ denote the coefficients of the
gradient and $V$ respectively.  
And,
\begin{align}
    V(f) = \sum_{j=1}^d \tilde{v}_j\frac{\partial}{\partial \tx_j} f\circ\varphi^{-1}.
\end{align}

Since $v$ is arbitrary, this means for all $j$
\begin{align}
    \sum_{i=1}^d \tilde{u}_i g_{ij} = \frac{\partial}{\partial \tx_j}f\circ\varphi^{-1}
    \quad \Longrightarrow \quad
    \tilde{u}_i = \sum_{j=1}^d g^{ij} \frac{\partial}{\partial \tx_j}f\circ\varphi^{-1}.
\end{align}

\paragraph{Riemannian divergence}
Recall that we define the Riemannian divergence using the patch coordinates in  (\ref{eq:riem-div-coord}), which we later show has a
coordinate-free form (\ref{eq:riem-div-intrinsic}) and can be computed in the ambient space (\ref{eq:riem-div-ambient}) if the manifold is embedded. 
The following theorem extends the Stokes theorem to Riemannian manifolds. 
\begin{mdframed}[style=MyFrame2]
\begin{restatable}[\textbf{Divergence theorem}]{thm}{div}
For any compactly supported $f\in \mathfrak{X}(\gM)$, $\int_\gM \nabla_g \cdot f
\,\ddd \mu_g = 0$.  
\end{restatable}
\end{mdframed}
\begin{proof}
Let $\{(\Psi_i, U_i)\}$ be a partition of unity. 
By compactness, we can choose a finite subcover over the support of $f$, so the
index set of $i$ is finite.  
\begin{align}
    \int_\gM \nabla_g \cdot f \,\ddd \mu_g 
    &= \int_\gM \nabla_g \cdot \left(\sum_{i} \Psi_i f\right) \,\ddd \mu_g \\
    &= \sum_{i} \int_{U_i} \nabla_g \cdot (\Psi_if) \,\ddd \mu_g \\
    &= \sum_{i} \int_{\varphi_i(U_i)} \nabla \cdot (\gvol\Psi_if)\circ\varphi^{-1} \,\ddd \tx.
\end{align}
All of the finitely many summands equal $0$ by an application of Stokes' theorem in
$\R^d$ \citep[Theorem 10.33]{rudin1976principles}. 
This is because the support of $\Psi_i\circ\varphi_i^{-1}$ is contained in
$\varphi_i(U_i))$; therefore at the boundary of $\varphi_i(U_i)$, $\Psi_i\circ\varphi_i^{-1}$ is equal to $0$.  

\end{proof}

The Riemannian divergence satisfies the following product rule.
\begin{mdframed}[style=MyFrame2]
\begin{restatable}[\textbf{Product rule}]{prop}{product}
\label{prop:product-rule}
Assume $V\in\mathfrak{X}(\gM)$ and $f\in C^\infty(\gM)$. 
Then 
\begin{align}
    \nabla_g\cdot (fV) = V(f) + f\nabla_g\cdot V.
    \label{eq:product-rule}
\end{align}
\end{restatable}
\end{mdframed}

\begin{proof}
Using (\ref{eq:riem-div-intrinsic}), the product rule of the Affine connection (see Appendix~\ref{app:covariant}),

\begin{align}
    \nabla_g \cdot (fV) &= \sum_{j=1}^d \langle \nabla_{\tilde{E}_j} (fV), \tilde{E}^j \rangle_g \\
    &= \sum_{j=1}^d \langle f \nabla_{\tilde{E}_j} V + \tilde{E}_j(f)V
    , \tilde{E}^j \rangle_g \\
    &= f\sum_{j=1}^d \langle \nabla_{\tilde{E}_j} V , \tilde{E}^j \rangle_g + \sum_{j=1}^d \tilde{E}_j(f)\left\langle \sum_{j'=1}^d \tilde{v}_{j'} \tilde{E}_{j'}
    , \tilde{E}^j \right\rangle_g  
    \\
    &= f\nabla_g \cdot V+ \sum_{j,j'=1}^d \tilde{E}_j(f)\tilde{v}_{j'}\langle  \tilde{E}_{j'}
    , \tilde{E}^j \rangle_g  
    \\
    &= f\nabla_g \cdot V+ \sum_{j,j'=1}^d \tilde{E}_j(f)\tilde{v}_{j'}\delta_{jj'}
    \\
    &= f\nabla_g \cdot V+ \sum_{j}^d \tilde{E}_j(f)\tilde{v}_{j}
    = f\nabla_g \cdot V+ V(f).
\end{align}

\end{proof}

\begin{mdframed}[style=MyFrame2]
\begin{restatable}[\textbf{Expanding Riemannian gradient}]{prop}{expand}
\label{prop:expand}
Let $V$ denote the tangential projection matrix in the sense of Proposition~\ref{prop:proj}.
Then for any $f\in C^\infty(\gM)$
\begin{align}
\sum_{k=1}^d V_k(f) V_k = \nabla_g f.
\end{align}
\end{restatable}
\end{mdframed}

\subsection{Covariant derivative}
\label{app:covariant}
An \textbf{affine connection}
allows us to compare values of a vector field at nearby points.
It is a differential operator denoted by
$\nabla : \mathfrak{X}(\gM)\times \mathfrak{X}(\gM)\rightarrow \mathfrak{X}(\gM)$ and written as $U,V\mapsto \nabla_U V$ for $U, V\in\mathfrak{X}(\gM)$, satisfying the following defining properties:
\begin{enumerate}
    \item Linearity in $U$: $\nabla_{fU_1 + gU_2} V = f\nabla_{U_1}V + g \nabla_{U_2} V$ for $f,g \in C^\infty(M)$ and $U_1,U_2,V\in\mathfrak{X}(M)$.
    \item Linearity in $V$: $\nabla_U(aV_1 + bV_2) = a\nabla_U V_1 + b\nabla_U V_2$ for $a,b\in\R$ and $U,V_1,V_2\in\mathfrak{X}(M)$.
    \item Product rule: $\nabla_U(fV) = f \nabla_UV + U(f)V$ for $f \in C^\infty(M)$ and $U, V\in\mathfrak{X}(M)$.
\end{enumerate}

$\nabla_U V$ is called the \textbf{covariant derivative} of $V$ in the $U$-direction. 

If $U,V\in\mathfrak{X}(\R^m)$, the Euclidean connection $\overbar{\nabla}$ is defined as 
\begin{align}
    \overbar{\nabla}_U V = \sum_{i=1}^m \sum_{j=1}^m \bar{u}_j \frac{\partial \bar{v}_i}{\partial x_j} \frac{\partial}{\partial x_i}.
\end{align}
It can be verified that the Euclidean connection is indeed an affine connection.

We can express a connection internally in terms of a coordinate system $\tilde{E}_i$. 
For any pair of indices $i$ and $j$, we define the connection coefficients of $\nabla$, denoted by $\Gamma$, as $d^3$ smooth functions satisfying
\begin{align}
    \nabla_{\tilde{E}_i}\tilde{E}_j = \sum_{k=1}^d \Gamma_{ij}^k \tilde{E}_k.
\end{align}

Then for any $U,V\in\mathfrak{X}(\gM)$, we have
\begin{align}
    \nabla_U V &= \nabla_U \sum_{j=1}^d \tilde{v}_j\tilde{E}_j  \\
    &= \sum_{j=1}^d \tilde{v}_j \nabla_U\tilde{E}_j + U(\tilde{v}_j)\tilde{E}_j \\
    &= \sum_{i,j=1}^d \tilde{u}_i \tilde{v}_j \nabla_{\tilde{E}_i}\tilde{E}_j + \sum_{j=1}^d U(\tilde{v}_j)\tilde{E}_j \\
    &= \sum_{i,j,k=1}^d \tilde{u}_i \tilde{v}_j \Gamma^k_{ij}\tilde{E}_k + \sum_{j=1}^d U(\tilde{v}_j)\tilde{E}_j.
    \label{eq:connection-coordinate}
\end{align}

Now given a metric tensor, we say that $\nabla$ is a \textbf{Levi-Civita connection of $g$} if it is 
\begin{enumerate}
    \item Compatible with $g$: $U(g(V, W)) = g(\nabla_U V, W) + g(V, \nabla_U W)$.
    \item Symmetric: $\nabla_u v - \nabla_V U = [U, V]$, where $[U,V] :=  \sum_{i=1}^d U(V_i) \tilde{E}_i - V(U_i) \tilde{E}_i$ is the Lie bracket.
\end{enumerate}

The first condition looks messy but it essentially says that the Levi-Civita connection leaves the metric invariant.  It is equivalent to saying that the covariant derivative of $g$ in any direction is zero.
\begin{mdframed}[style=MyFrame2]
\begin{restatable}[\textbf{Fundamental Theorem of Riemannian Geometry}]{thm}{funda}
Let $(\mathcal{M},g)$ be a Riemannian manifold. 
There exists a unique Levi-Civita connection of $g$. 
\end{restatable}
\end{mdframed}
See \citet[Theorem 5.10]{lee2018riemannian} for proof.
The connection coefficients of the Levi-Civita connection are called the \textbf{Christoffel symbols} of $g$. 
They are symmetric in the lower indices, \ie $\Gamma^k_{ij}= \Gamma^k_{ji}$.
A by-product of the proof of the fundamental theorem is the following identity, which will turn out to be useful in deriving the identity for the Riemannian divergence:
\begin{align}
    \frac{\partial}{\partial \tx_j} g_{ki} = \sum_{l=1}^d \Gamma^l_{jk} g_{li} + \Gamma^l_{ji} g_{lk}.
    \label{eq:christoffel-identity}
\end{align}

An example of a Levi-Civita connection is the Euclidean connection of $(\R^d, \bar{g})$. 
It can be checked that $\overbar{\nabla}$ is both symmetric and compatible with $\bar{g}$.
Furthermore, for any $d$-submanifold $\gM$ embedded in $\R^m$ for $m>d$, we can define a \textbf{tangential connection}
\begin{align}
    \nabla^\top_UV = P\overbar{\nabla}_{\overbar{U}} \overbar{V}
\end{align}
for $U,V\in\mathfrak{X}(\gM)$, where $\overbar{U}$ and $\overbar{V}$ are any\footnote{The value of the tangential connection is independent of the extensions chosen, so $\nabla^\top$ is well-defined.} smooth extensions of $U$ and $V$ to $\R^m$.
$P$ is the tangential projection defined as
\begin{align}
    (PV)(x) = \sum_{j=1}^m (P_x \bar{v})_j \frac{\partial }{\partial x_j}
\end{align}
for any $V\in \mathfrak{X}(\R^m)$.
Recall that $P_x$ is the \textbf{orthogonal projection} onto the tangent space spanned by $\frac{\partial \psi}{\partial \tx_i}$. 
\emph{The tangential connection $\nabla^\top$ is the Levi-Civita connection on the embedded submanifold $\gM$} \citep[Proposition 5.12]{lee2018riemannian}.

\newpage
\section{Proofs}

\begin{mdframed}[style=MyFrame2]
\marginal*
\end{mdframed}

\begin{proof}
Our first step is to express the time derivative of the density using \textit{derivations} (spatial derivatives); this gives us a partial differential equation (PDE) on the manifold.  
Second, we apply the Feynman-Kac formula \citep[Proposition 3.1]{sde2021thalmaier} to the solution of the PDE.

We denote by $\ddd \tilde{X_t} = \tilde{v}_0 \dt + \tilde{v} \circ \ddd B_t$ the Stratonovich SDE defined on the patch.
The density $p$ of the process satisfies the Fokker-Planck equation \citep[Equation (8.16)]{chirikjian2009stochastic}:
\begin{align}
    \partt{p(\tilde{x}, t)} = \underbrace{-\igvol \diverge (\gvol \tilde{v}_0 p)}_{\text{first term}}+ 
    \underbrace{
    \frac{1}{2} \igvol 
    \sum_{i=1}^{d}
    \sum_{j=1}^{d} 
        \frac{\partial}{\partial \tilde{x}_{i}} 
            \Big( 
                \sum_{k=1}^{w}
                    \tilde{v}_{i,k} 
                        \frac{\partial}{\partial \tilde{x}_{j}} 
                            \big(
                                \gvol \tilde{v}_{j,k} p
                            \big)
            \Big) 
    }_{\text{second term}}
\label{eq:time-derivative}
\end{align}
We would like to re-express the RHS using the abstract vectors $V_0$ and $V$.
Note the first term can be written as $-\nabla_g \cdot (pV_{0})$. 
We now show that we can also rewrite the second term in terms of the Riemannian divergence.
\begin{align}
& \frac{1}{2} \igvol 
    \sum_{i=1}^{d}
    \sum_{j=1}^{d} 
        \frac{\partial}{\partial \tilde{x}_{i}} 
            \Big( 
                \sum_{k=1}^{w}
                    \tilde{v}_{i,k} 
                        \frac{\partial}{\partial \tilde{x}_{j}} 
                            \big(
                                \gvol \tilde{v}_{j,k} p
                            \big)
            \Big) 
\\&
= \frac{1}{2} \igvol 
    \sum_{k=1}^{w}
    \sum_{i=1}^{d}
        \frac{\partial}{\partial \tilde{x}_{i}} 
            \Big( 
                    \tilde{v}_{i,k} 
                    \sum_{j=1}^{d} 
                        \frac{\partial}{\partial \tilde{x}_{j}} 
                            \big(
                                \gvol \tilde{v}_{j,k} p
                            \big)
            \Big) 
\\&
= \frac{1}{2} \igvol 
    \sum_{k=1}^{w}
    \sum_{i=1}^{d}
        \frac{\partial}{\partial \tilde{x}_{i}} 
            \Big( 
                    \tilde{v}_{i,k} 
                    \gvol
                    \textcolor{violet}
                    {\igvol
                    \sum_{j=1}^{d} 
                        \frac{\partial}{\partial \tilde{x}_{j}} 
                            \big(
                                \gvol \tilde{v}_{j,k} p
                            \big)}
            \Big) 
\\&
= \frac{1}{2} \igvol 
    \sum_{k=1}^{w}
    \sum_{i=1}^{d}
        \frac{\partial}{\partial \tilde{x}_{i}} 
            \Big( 
                    \tilde{v}_{i,k} 
                    \gvol
                    \textcolor{violet}{\nabla_g \cdot (pV_k)}
            \Big) 
\\&
=   \frac{1}{2}   
    \sum_{k=1}^{w}
    \textcolor{cyan}
     {\igvol 
    \sum_{i=1}^{d}
        \frac{\partial}{\partial \tilde{x}_{i}} 
            \Big( 
                    \gvol
                    \tilde{v}_{i,k} 
                    \nabla_g \cdot (pV_k)
            \Big) }
\\&
=   \frac{1}{2}
    \sum_{k=1}^{w}
    \textcolor{cyan}{
    \nabla_g \cdot
    \Bigg(
        \Big( 
                \nabla_g \cdot (pV_k)
        \Big)
         V_k
    \Bigg)}
\end{align}
Summing these two terms give us
\begin{align}
\partt{p(x, t)} = -\nabla_g \cdot (pV_{0}) + 
\frac{1}{2}
    \sum_{k=1}^{w}
    \nabla_g \cdot
    \Bigg(
        \Big( 
                \nabla_g \cdot (pV_k)
        \Big)
         V_k
    \Bigg)
\label{eq:td-in-rd} %
\end{align}
Next, we expand the above formula using the product rule (\ref{eq:product-rule}):

\begin{align}
&
\partt{p(x, t)} = \textcolor{violet}{-\nabla_g \cdot (pV_{0})}
+ 
\frac{1}{2}
    \sum_{k=1}^{w}
    \nabla_g \cdot
    \Bigg(
        \Big( 
                \textcolor{blue}{\nabla_g \cdot (pV_k)}
        \Big)
         V_k
    \Bigg)
\\&
= \textcolor{violet}{-V_0(p) -p\nabla_g\cdot(V_0)}
+ 
\frac{1}{2}
    \sum_{k=1}^{w}
    \nabla_g \cdot
    \Bigg(
        \Big( 
               \textcolor{blue}{V_k(p) + p \nabla_g\cdot(V_k)}
        \Big)
         V_k
    \Bigg)
    \label{eq:rdexpansion1}
\\&
= -V_0(p) -p\nabla_g\cdot(V_0)
+ 
\frac{1}{2}
    \sum_{k=1}^{w}
    \textcolor{red}{
    \nabla_g \cdot
    \Bigg(
        \Big( 
              V_k(p) + p \nabla_g\cdot(V_k)
        \Big)
         V_k
    \Bigg)}
\\&
= -V_0(p) -p\nabla_g\cdot(V_0)
+ 
\frac{1}{2}
    \sum_{k=1}^{w}
    \textcolor{red}{
    \Bigg(
     V_k
        \Big( 
              V_k(p) + p \nabla_g\cdot(V_k)
        \Big)
    +
    \Big( 
              V_k(p) + p \nabla_g\cdot(V_k)
        \Big)
    \nabla_g \cdot (V_k)
    \Bigg)}
\\&
=  -V_0(p) -p\nabla_g\cdot(V_0)
+ 
\frac{1}{2}
    \sum_{k=1}^{w}
    \Bigg(
     V_k(V_k(p)) + V_k(p \nabla_g\cdot(V_k))
    +
    V_k(p) \nabla_g \cdot (V_k) + \textcolor{orange}{p \nabla_g\cdot(V_k) \nabla_g \cdot (V_k)}
    \Bigg)
\\&
=  -V_0(p) -p\nabla_g\cdot(V_0)
+ 
\frac{1}{2}
    \sum_{k=1}^{w}
    \Bigg(
     \textcolor{purple}{V_k(V_k(p))} + V_k(p \nabla_g\cdot(V_k))
    +
    V_k(p) \nabla_g \cdot (V_k) + \textcolor{orange}{p (\nabla_g\cdot(V_k))^2}
    \Bigg)
\\&
=  -V_0(p) -p\nabla_g\cdot(V_0)
+ 
\frac{1}{2}
    \sum_{k=1}^{w}
    \Bigg(
     \textcolor{purple}{V_k^2(p)} +
     \textcolor{teal}{V_k(p \nabla_g\cdot(V_k))}
    +
    V_k(p) \nabla_g \cdot (V_k) + p (\nabla_g\cdot(V_k))^2
    \Bigg) 
\\&
=  -V_0(p) -p\nabla_g\cdot(V_0)
+ 
\frac{1}{2}
    \sum_{k=1}^{w}
    \Bigg(
     V_k^2(p) +
     \textcolor{teal}{V_k(p) \nabla_g\cdot(V_k) + pV_k(\nabla_g\cdot(V_k))}
    +
    V_k(p) \nabla_g \cdot (V_k) + p (\nabla_g\cdot(V_k))^2
    \Bigg)
\\&
=  -V_0(p) -p\nabla_g\cdot(V_0)
+ 
\frac{1}{2}
    \sum_{k=1}^{w}
    \Bigg(
     V_k^2(p) +
     \textcolor{blue}{V_k(p) \nabla_g\cdot(V_k)} + pV_k(\nabla_g\cdot(V_k))
    +
    \textcolor{blue}{V_k(p) \nabla_g \cdot (V_k)} + p (\nabla_g\cdot(V_k))^2
    \Bigg)
\\&
=  -V_0(p) -p\nabla_g\cdot(V_0)
+ \textcolor{blue}{ \sum_{k=1}^{w} V_k(p) \nabla_g\cdot(V_k)} + 
\frac{1}{2}
    \sum_{k=1}^{w}
    \Bigg(
     V_k^2(p) +
     pV_k(\nabla_g\cdot(V_k))
     + p (\nabla_g\cdot(V_k))^2
     \Bigg)
\\&
=  -V_0(p) -p\nabla_g\cdot(V_0)
+ \textcolor{blue}{ \sum_{k=1}^{w} \big( V_k \nabla_g\cdot(V_k) \big) (p)} + 
\frac{1}{2}
    \sum_{k=1}^{w}
    \Bigg(
     V_k^2(p) +
     pV_k(\nabla_g\cdot(V_k))
     + p (\nabla_g\cdot(V_k))^2
     \Bigg)
\\&
=  -V_0(p) -p\nabla_g\cdot(V_0)
+ \textcolor{blue}{\big( (V \cdot \nabla_g )  V \big) (p)} + 
\frac{1}{2}
    \sum_{k=1}^{w}
    \Bigg(
     V_k^2(p) +
     \textcolor{olive}{pV_k(\nabla_g\cdot(V_k))
     + p (\nabla_g\cdot(V_k))^2}
     \Bigg)
\\&
=  -V_0(p) -p\nabla_g\cdot(V_0)
+ \big( (V \cdot \nabla_g )  V \big) (p) + 
\frac{1}{2}
    \sum_{k=1}^{w}
    \Bigg(
     V_k^2(p) +
     \textcolor{olive}{p \nabla_g \cdot ((\nabla_g \cdot V_k) V_k )}
     \Bigg)
\\&
=  -V_0(p) -p\nabla_g\cdot(V_0)
+ \big( (V \cdot \nabla_g )  V \big) (p) + 
\frac{1}{2}
    \sum_{k=1}^{w}
    \Bigg(
     V_k^2(p) +
     \textcolor{olive}{p \nabla_g \cdot ((\nabla_g \cdot V_k) V_k)}
     \Bigg)
\\&
= -V_0(p) -p\nabla_g\cdot(V_0)
+ \big( (V \cdot \nabla_g )  V \big) (p) + 
\frac{1}{2}
\sum_{k=1}^{w}
V_k^2(p) +
\textcolor{magenta}{
\frac{1}{2}
\sum_{k=1}^{w}
p \nabla_g \cdot ((\nabla_g \cdot V_k) V_k)
}
\\&
= -V_0(p) -p\nabla_g\cdot(V_0)
+ \big( (V \cdot \nabla_g )  V \big) (p) + 
\frac{1}{2}
\sum_{k=1}^{w}
V_k^2(p) +
\textcolor{magenta}{
\frac{1}{2}
p \nabla_g \cdot ((V \cdot \nabla_g) V)
}
\label{eq:expanded-rd}
\end{align}
In order to apply the Feynman-Kac formula, we group all the terms by the order of differentiation (of $p$), which gives us
\begin{align}
\partt{p(x, t)} 
&
=  \textcolor{blue}{-V_0}(p) -p\textcolor{violet}{\nabla_g\cdot(V_0)}
+ \textcolor{blue}{\big( (V \cdot \nabla_g )  V \big)} (p) + 
\textcolor{olive}{\frac{1}{2}
\sum_{k=1}^{w}
 V_k^2}(p)
 + 
 \textcolor{violet}{\frac{1}{2}}
  p \textcolor{violet}{\nabla_g \cdot ((V \cdot \nabla_g) V)}
\\&
= p \underbrace{\bigg(\textcolor{violet}{-\nabla_g\cdot(V_0)+\frac{1}{2} \nabla_g \cdot ((V \cdot \nabla_g) V) )} \bigg)}_{\textcolor{violet}{\cV}}
 +\bigg( \textcolor{blue}{ -V_0+\big( (V \cdot \nabla_g )  V \big)} \bigg)(p)
 + \bigg( \textcolor{olive}{\frac{1}{2} \sum_{k=1}^{w} V_k^2} \bigg) (p)
\end{align}

Now the above is a parabolic PDE, which can be solved using the Feynman-Kac formula \citep[Proposition 3.1]{sde2021thalmaier}. 
Let $Y$ be induced (\ref{eq:fk-sde}) restated below
\begin{equation}
\left\{\begin{array}{l}
\ddd Y= ( \textcolor{blue}{-V_0 + (V\cdot \nabla_g)  V}) d t+ \textcolor{olive}{\sum_{k=1}^{w} (V_k)} \circ \ddd {B'_s}^k \\
Y_{0}=x \\
\end{array}\right.
\end{equation}
Then $p(x, t)$ is given by
\begin{equation}
p(x, t)=\mathbb{E}\left[\exp \left(\int_{0}^{t} \textcolor{violet}{\mathcal{V}}\left(Y_{s}(x)\right) d s\right) p_0\left(Y_{t}\right)\, \Bigg\vert \, \vy_0=x\right]
\label{eq:pxt}
\end{equation}
where $p_0 = p(x, 0)$ is the prior distribution.

\end{proof}

\newpage
\begin{mdframed}[style=MyFrame2]
\rctelbo*
\end{mdframed}

\begin{proof}

Let $\sP$ be the probability measure under which $B'$ is a Brownian motion. 
Let 
\begin{align}
    \ddd\hat{B} = - a \ds + \ddd B'_s,
    \label{eq:shifted-bm}
\end{align}
where $a$ is the variational degree of freedom. 
Let $\sQ$ be defined as
\begin{equation}
    {\ddd \sQ} = \exp \left(\int_{0}^{T} a(Y_s, s) \ddd B'_s -\frac{1}{2} \int_{0}^{T}\|a(Y_s, s)\|_{2}^{2} \ds\right) {\ddd \sP}.
    \label{dqoverdp}
\end{equation}
Note that the first term is an It\^o integral. 
Then by the Girsanov theorem \citep[Theorem 8.6.3]{oksendal2003stochastic}, 
$\hat{B}$ is a Brownian motion wrt $\sQ$.
Therefore, changing the measure from $\sP$ to $\sQ$ to the expression in Theorem~\ref{thm:marginal-density} yields
\begin{align*}
\log p(x, t)=\log \mathbb{E}_\sQ\left[ \frac{\ddd\sP}{\ddd\sQ}\cdot p_0\left(Y_{t}\right) \exp \left(- \int_{0}^{T} 
\nabla_g\cdot \left(V_0 - \frac{1}{2}  (V \cdot \nabla_g) V\right)
d s\right) \, \Bigg\vert \, \vy_0=x \right],
\label{eq:exactlogp_in_thm2}
\end{align*}
which by Jensen's inequality, is lower bounded by
\begin{align}
\mathbb{E}_\sQ\left[ \log \frac{\ddd\sP}{\ddd\sQ} + \log p_0\left(Y_{t}\right) -\left(\int_{0}^{T} 
\nabla_g\cdot \left(V_0 + \frac{1}{2}  (V \cdot \nabla_g) V\right)
d s\right) \, \Bigg\vert \, \vy_0=x \right].
\end{align}
Now under the expectation, the Radon-Nikodym derivative can be simplified:
\begin{align}
    \E_\sQ\left[\log \frac{\ddd \sP}{\ddd \sQ}  \, \Bigg\vert \, \vy_0=x \right]
    &= \E_\sQ\left[
    - \int_{0}^{T} a(Y_s, s) \ddd B'_s + \frac{1}{2} \int_{0}^{T}\norm{a(Y_s, s)}_{2}^{2} \ds  \, \Bigg\vert \, \vy_0=x 
    \right] \\
    &= \E_\sQ\left[
    -\int_{0}^{T} a(Y_s, s) \ddd \hat{B}_s - \frac{1}{2} \int_{0}^{T}\norm{a(Y_s, s)}_{2}^{2} \ds  \, \Bigg\vert \, \vy_0=x 
    \right] \\
    &= \E_\sQ\left[
    - \frac{1}{2} \int_{0}^{T}\norm{a(Y_s, s)}_{2}^{2} \ds  \, \Bigg\vert \, \vy_0=x 
    \right] 
\end{align}
where we used the definition of $\sQ$ (\ref{dqoverdp}), the definition of $\ddd \hat{B}$ (\ref{eq:shifted-bm}), and the Martingale property of the It\^o integral \citep[Corollary 3.2.6]{oksendal2003stochastic}. 
This concludes the proof.

\end{proof}

\newpage
\begin{mdframed}[style=MyFrame2]
\divergence*
\end{mdframed}

\begin{proof}
We drop the index on $k$ (since the statement is for any smooth vector). 
Using product rule, the LHS of (\ref{eq:riem-div-intrinsic}) is equal to
\begin{align}
    \sum_{j=1}^d \frac{\partial \tilde{v}_j}{\partial \tx_j} + \tilde{v}_j \igvol \frac{\partial}{\partial \tx_j} \gvol
\end{align}
Using the chain rule, Jacobi's formula, and the identity (\ref{eq:christoffel-identity}), we have
\begin{align}
    \tilde{v}_j \igvol \frac{\partial}{\partial \tx_j} \gvol 
    &= \frac{1}{2}\tilde{v}_j |G|^{-1} \frac{\partial}{\partial \tx_j} \det G \\
    &= \frac{1}{2}\tilde{v}_j \tr\left(G^{-1}\frac{\partial G}{\partial \tx_j}\right) \\
    &= \frac{1}{2}\tilde{v}_j \sum_{i,k=1}^d g^{ik} \frac{\partial g_{ki}}{\partial\tx_j} \\
    &= \frac{1}{2}\tilde{v}_j \sum_{i,k=1}^d g^{ik} \left(\sum_{l=1}^d \Gamma^l_{jk} g_{li} + \Gamma^l_{ji} g_{lk}\right) \\
    &= \frac{1}{2}\tilde{v}_j \left(\sum_{i,k,l=1}^d \Gamma^l_{jk} g^{ik} g_{li} + 
    \tilde{v}_j \sum_{i,k,l=1}^d \Gamma^l_{ji} g^{ik} g_{lk}\right) \\
    &= \frac{1}{2}\tilde{v}_j \sum_{k,l=1}^d \Gamma^l_{jk} \delta_{kl} + 
    \frac12\tilde{v}_j \sum_{i,l=1}^d \Gamma^l_{ji} \delta{il} \\
    &= \frac{1}{2}\tilde{v}_j \sum_{k=1}^d \Gamma^k_{jk} + \frac12
    \tilde{v}_j \sum_{i=1}^d \Gamma^i_{ji} \\
    &= \tilde{v}_j \sum_{k=1}^d \Gamma^k_{jk} 
\end{align}

Therefore, the LHS reduces to 
\begin{align}
    \sum_{j=1}^d \left(\frac{\partial \tilde{v}_j}{\partial \tx_j} + \tilde{v}_j \sum_{k=1}^d \Gamma^k_{jk} \right)
\end{align}

Now we express the covariant derivative on the RHS using the connection coefficients (\ref{eq:connection-coordinate})
\begin{align}
    \nabla_{\tilde{E}_j} V
    = \sum_{i,k=1}^d \tilde{v}_i \Gamma^k_{ji} \tilde{E}_k + \sum_{i=1}^d \frac{\partial \tilde{v}_i}{\partial \tx_j} \tilde{E}_i
\end{align}
which means
\begin{align}
    \langle\nabla_{\tilde{E}_j} V, \tilde{E}^j\rangle_g
    &= \sum_{i,k=1}^d \tilde{v}_i \Gamma^k_{ji} \langle \tilde{E}_k, \tilde{E}^j\rangle_g + \sum_{i=1}^d \frac{\partial \tilde{v}_i}{\partial \tx_j} \langle \tilde{E}_i, \tilde{E}^j\rangle_g \\
    &= \sum_{i,k=1}^d \tilde{v}_i \Gamma^k_{ji} \delta_{kj} + \sum_{i=1}^d \frac{\partial \tilde{v}_i}{\partial \tx_j} \delta_{ij} \\
    &= \sum_{i=1}^d \tilde{v}_i \Gamma^j_{ji} + \frac{\partial \tilde{v}_j}{\partial \tx_j}.
\end{align}
Now summing all terms yields
\begin{align}
    \sum_{j=1}^d\langle\nabla_{\tilde{E}_j} V, \tilde{E}^j\rangle_g =
    \sum_{i,j=1}^d \tilde{v}_i  \Gamma^j_{ji} + \sum_{j=1}^d\frac{\partial \tilde{v}_j}{\partial \tx_j}.
\end{align}
Relabeling $i\rightarrow j$ and $j\rightarrow k$ in the term term shows this is equal to the LHS. 

For the second half of the theorem, recall that the Levi-Civita connection is equal to the tangential connection. 
Therefore, changing the basis via $\tilde{E}_j = \sum_{k=1}^m \frac{\partial\psi_k}{\partial \tilde{x}_j} \frac{\partial}{\partial x_k}$, we can rewrite it as 
\begin{align}
    (\nabla_{\tilde{E}_j} V)(x)
    = \left(P \left(
    \sum_{i=1}^m \sum_{k=1}^m \frac{\partial\psi_k}{\partial \tilde{x}_j} \frac{\partial \bar{v}_i}{\partial x_k} \frac{\partial}{\partial x_i}\right)\right)(x)
    = \sum_{i=1}^m \left(P_x \frac{\ddd \bar{v}}{\ddd x} \frac{\ddd \psi}{\ddd \tx}\right)_{ij} \frac{\partial}{\partial x_i}.
\end{align}

On the other hand,
\begin{align}
    \tilde{E}^j 
    = \sum_{k=1}^d g^{kj}\tilde{E}_k 
    = \sum_{i=1}^m \sum_{k=1}^d \left(\frac{\ddd \psi}{\ddd \tx}^\top\frac{\ddd\psi}{\ddd \tx}\right)^{-1}_{kj} 
    \frac{\partial\psi_i}{\partial \tilde{x}_k} \frac{\partial}{\partial x_i}
    = \sum_{i=1}^m \left(\frac{\ddd\psi}{\ddd\tx}\left(\frac{\ddd \psi}{\ddd \tx}^\top \frac{\ddd\psi}{\ddd\tx}\right)^{-1}\right)_{ij} \frac{\partial}{\partial x_i}.
\end{align}

Since $g$ is the induced metric, the summation over $j=1,\cdots,d$ is equivalent to the Frobenius inner product $\langle\cdot,\cdot\rangle_F$ of the two $m\times d$ matrices
\begin{align}
    \sum_{j=1}^d \langle \nabla_{\tilde{E}_j}V, \tilde{E}^j \rangle_g 
    &= \left\langle P_x \frac{\ddd \bar{v}}{\ddd x} \frac{\ddd \psi}{\ddd \tx},    \frac{\ddd\psi}{\ddd\tx}\left(\frac{\ddd \psi}{\ddd \tx}^\top \frac{\ddd\psi}{\ddd\tx}\right)^{-1}
    \right\rangle_F \\
    &= \tr\left(P_x \frac{\ddd \bar{v}}{\ddd x} \frac{\ddd \psi}{\ddd \tx} \left(\frac{\ddd \psi}{\ddd \tx}^\top \frac{\ddd\psi}{\ddd\tx}\right)^{-1} \frac{\ddd\psi}{\ddd\tx}^\top\right) \\
    &= \tr\left(P_x \frac{\ddd \bar{v}}{\ddd x} P_x\right).
\end{align}

\end{proof}

\begin{mdframed}[style=MyFrame2]
\projection*
\end{mdframed}

\begin{proof}
By definition,
\begin{align}
    (V\cdot \nabla_g) V = \sum_{j=1}^m V_j \nabla_g \cdot V_j.
\end{align}

Denote by the $j$'th column of $P_x$ by $(P_x)_{:j}$.
Applying the resulting tangent vector to any smooth function $f$ (evaluated at $x$) and applying (\ref{eq:riem-div-ambient}) gives
\begin{align}
    ((V\cdot \nabla_g) V)(f)(x) 
    &= \sum_{j=1}^m (\nabla_g \cdot V_j)(x) V_j(f)(x) \\
    &= \sum_{j=1}^m \tr\left(P_x \frac{\ddd {(P_x)_{:j}}}{\ddd x} P_x\right) \sum_{i=1}^m (P_x)_{ij} \frac{\partial f}{\partial x_i} \\
    &=  \sum_{i=1}^m \sum_{j=1}^m (P_x)_{ij} \tr\left(P_x \frac{\ddd {(P_x)_{:j}}}{\ddd x} P_x\right)  \frac{\partial f}{\partial x_i}. 
\end{align}
That is, the resulting tangent vector's coefficients correspond to the tangential projection of the vector
\begin{align}
    \begin{bmatrix}
    \tr\left(P_x \frac{\ddd {(P_x)_{:1}}}{\ddd x} P_x\right) \\
    \vdots \\
    \tr\left(P_x \frac{\ddd {(P_x)_{:m}}}{\ddd x} P_x\right)
    \end{bmatrix}
    \label{eq:div_proj_vector}
\end{align}
which we claim is orthogonal to the tangential linear subspace. 

To prove the claim, we first note that we can rewrite $P_x$ as
\begin{align}
    P_x = I - n_x n_x^\top
\end{align}
where $n_x$ is of type $\R^{m\times (m-d)}$, and the column vectors of $n_x$ are orthonormal, and orthogonal to the tangential linear subspace; that is to say, $P_x n_x = 0$. 
Using this representation, we can write the Jacobian as
\begin{align}
    \left(\frac{\ddd (P_x)_{:j}}{\ddd x}\right)_{kl}
    &= - \sum_{r=1}^{m-d} \frac{\partial}{\partial x_l} (n_x)_{kr}(n_x)_{jr} \\
    &= - \sum_{r=1}^{m-d} (n_x)_{jr}\frac{\partial}{\partial x_l} (n_x)_{kr} + (n_x)_{kr}\frac{\partial}{\partial x_l} (n_x)_{jr}.
\end{align}

Now multiplying by the projection matrix from both sides gives

\begin{align}
    P_x\frac{\ddd (P_x)_{:j}}{\ddd x} P_x
    &= - \sum_{r=1}^{m-d} (n_x)_{jr}
    P_x
    \begin{bmatrix}
    \nabla_x (n_x)_{1r}^\top \\
    \vdots \\
    \nabla_x (n_x)_{mr}^\top
    \end{bmatrix}
    P_x
    + \underbrace{P_x(n_x)_{:r}}_{0} \nabla_x (n_x)_{jr}^\top P_x.
\end{align}

Lastly, let 
\begin{align}
    \tau_r = \tr\left(
    P_x
    \begin{bmatrix}
    \nabla_x (n_x)_{1r}^\top \\
    \vdots \\
    \nabla_x (n_x)_{mr}^\top
    \end{bmatrix}
    P_x
    \right)
\end{align}
which means (\ref{eq:div_proj_vector}) is simply
\begin{align}
    -\sum_{r=1}^{m-d} (n_x)_{:r} \tau_r.
\end{align}

This implies the claim is true, since this is nothing more than a linear combination of the column vectors of $n_x$, which is orthogonal to the tangential linear subspace. 

\end{proof}

\begin{mdframed}[style=MyFrame2]
\equivalent*
\end{mdframed}

\begin{proof}

We work with the derivation version of (\ref{eq:fixed-inference}):
\begin{align}
    \ddd Y = U_0\dt + V \circ \ddd \hat{B}_s,
    \label{eq:fixed-inference-derivation}
\end{align}
That is, $U_0(f) = \sum_k (Pr)_k \frac{\partial}{\partial \tx_k} f\circ \psi$, and $V$ is the tangential projection.
The marginal density $q$ follows the Fokker-Planck PDE
\begin{align}
    \parts{q} &= -\nabla_g \cdot (qU_{0}) + \frac{1}{2}\sum_{k=1}^m\nabla_g \cdot\left(\left(\nabla_g \cdot (qV_k)\right)V_k\right) \\
    &= -\nabla_g \cdot (qU_{0}) + \frac{1}{2}\sum_{k=1}^m\nabla_g \cdot\left(\left(
    V_k(q) + q \nabla \cdot V_k
    \right)V_k\right)\\
    &= -\nabla_g \cdot (qU_{0}) + \frac{1}{2}\sum_{k=1}^m\nabla_g \cdot\left(
    V_k(q) V_k
    \right)\\
    &= -\nabla_g \cdot (qU_{0}) + \frac{1}{2}\sum_{k=1}^m\nabla_g \cdot\left(
    qV_k(\log q) V_k
    \right)\\
    &= -\nabla_g \cdot (qU_{0}) + \frac{1}{2}\nabla_g \cdot\left(
    q\nabla_g \log q
    \right),
\end{align}
where we have used the product rule, and Proposition~\ref{prop:proj}, the chain rule, and Proposition~\ref{prop:expand}.

For $\lambda\leq1$, we can rearrange the Fokker-Planck and get
\begin{align}
    \parts{q} =  - \nabla_g\cdot \left( q\left(U_0 - \frac{\lambda}{2} \nabla_g\log q \right)\right) + \frac{1-\lambda}{2}\nabla_g\cdot\left(q\nabla_g \log q\right),
    \label{eq:fp-equiv}
\end{align}
which is the Fokker-Planck equation of the process (\ref{eq:equiv-lambda}).

To construct a reverse process inducing the same family of marginal densities, we mirror the diffusion term around $0$:

\begin{align}
    \parts{q} = - \nabla_g\cdot \left( q\left(U_0 - \left(1-\frac{\lambda}{2}\right)\nabla_g \log q\right) \right) - \frac{1-\lambda}{2}\nabla_g\cdot\left(q\nabla_g \log q\right)
    \label{eq:fp-equiv-forward}
\end{align}
Now we apply a change of variable of time via $p(x,t) = q(x,T-t)$,
which means $\partt p = - \parts q |_{s=T-t}$ and thus
\begin{align}
    \parts{p} = - \nabla_g\cdot \left( q\left( \left(1-\frac{\lambda}{2}\right)\nabla_g \log q - U_0 \right)\right) + \frac{1-\lambda}{2}\nabla_g\cdot\left(q\nabla_g \log q\right),
    \label{eq:fp-equiv-forward-time-changed}
\end{align}
which is the Fokker-Planck of (\ref{eq:equiv-lambda-reverse}).

\end{proof}

\begin{mdframed}[style=MyFrame2]\
\score*
\end{mdframed}

\begin{proof}

Approximating $\nabla_g \log q$ in (\ref{eq:equiv-lambda-reverse}) using $S_\theta$ and plugging it in (\ref{eq:genmodel}) and (\ref{eq:equiv-lambda}) in (\ref{eq:inference-sde}), we get
\begin{align}
    V_0 &= \left(1-\frac{\lambda}{2}\right)S_\theta - U_0 \label{eq:equiv-lambda-reverse-plugin}\\
    \sqrt{1-\lambda}Va
    &= \left(1- \lambda\right)S_\theta + \frac{\lambda}{2} \left( S_\theta - \nabla_g \log q \right). \label{eq:lambda-induced-a}
\end{align}

Also, as we only need to focus on the tangential components of $a$, note that
\begin{align}
    \norm{V a}_g^2 
    &= \left\langle \sum_k V_k a_k, \sum_{k'} V_{k'} a_{k'} \right\rangle_g \\
    &= \sum_{kk'} a_k a_{k'}\left\langle  V_k, V_{k'}  \right\rangle_g \\
    &= \sum_{kk'} a_k a_{k'}\left\langle \sum_j P_{jk}{E}_{j}, \sum_{j'} P_{j'k'} {E}_{j'} \right\rangle_g \\
    &= \sum_{kk'jj'} a_k a_{k'}P_{jk}P_{j'k'}\left\langle  {E}_{j}, {E}_{j'}  \right\rangle_g \\
    &= \sum_{kk'j} a_k a_{k'}P_{jk}P_{jk'} = \norm{Pa}_2^2,
\end{align}
where $E_j$ denote the ambient space Euclidean derivation $\frac{\partial}{\partial x_j}$.

Thus, we have
\begin{align*}
    \frac{1}{2} \norm{Pa}_{2}^2
    &= \frac{1}{2(1-\lambda)} \left[ \left( 1 - \lambda \right)^2 \norm{S_\theta}_{g}^2 + \left( 1 - \lambda \right) \lambda \langle S_\theta, S_\theta - \nabla_g \log q \rangle_g + \frac{\lambda^2}{4} \norm{ S_\theta - \nabla_g \log q }_{g}^2 \right] \\
    &= \left( 1 - \frac{\lambda}{2}\right) \frac{1}{2}\norm{S_\theta}_{g}^2 + \frac{\lambda}{2} \left( \frac{1}{2}\norm{S_\theta}_{g}^2 -  \langle S_\theta, \nabla_g \log q \rangle_g \right) + \frac{\lambda^2}{4(1-\lambda)}\frac{1}{2}\norm{S_\theta - \nabla_g \log q}_{g}^2 %
\end{align*}
\begin{gather*}
    \nabla \cdot V_0 
    =  \left(1 - \frac{\lambda}{2} \right) \nabla \cdot \left(S_\theta - \left(\frac{2}{2-\lambda}\right)U_0\right)
\end{gather*}
Summing up these two parts gives us $\gE_\lambda^\infty$.
Taking the expectation over $q(\cdot, 0)$ and applying the divergence theorem give us the desired identity. 
\end{proof}

\section{Manifolds}
\label{app:manifolds}
\everypar{\looseness=-1}
We provide some background on the manifolds used in this paper. 

\subsection{Spheres and tori}
\label{app:spheres}
Spheres are defined as submanifolds in an Euclidean space of points with unit Euclidean norm. 
Precisely, an $d$-sphere is $\sS^d = \{x\in \R^{d+1}: \norm{x}_2 =1 \}$. 
Therefore the ambient space dimensionality of a $d$ sphere is $m=d+1$. 
Tori are products of $1$-spheres (or circles); that is $\sT^d = \Pi_{i=1}^d \sS^1 $.
Naturally, we can embed a $d$-torus in a $m=2d$-dimensional ambient space. 

\paragraph{Tangential projection}
Without loss of generality, we derive the orthogonal projection to the tangent space of spheres. 
The tangential projection of tori is just the same linear operator applied to $d$ $\R^2$ vectors independently.

To derive the tangential project, we note that any any incremental change in $x$, denoted by $\ddd x$, will need to leave the norm $\norm{x}_2$ unchanged. 
That is,
\begin{align}
    \ddd \norm{x}_2^2 = 2 x \,\ddd x = 0.
\end{align}

This means $x$ is normal to the tangential linear subspace.
We can find the orthogonal projection onto the tangent space by subtracting the normal component, via $P_x= I - \frac{xx^\top}{\norm{x}_2^2}$.

\paragraph{Closest-point projection}
The closest-point projection onto the sphere is $\pi(x) = \frac{x}{\norm{x}_2}$. 
One can verify this is the point on $\sS^d$ that minimizes the Euclidean distance from $x\in\R^{d+1}\setminus\{0\}$. 

\subsection{Hyperbolic spaces}
\label{app:hyperbolics}
We work with the Lorentzian model of the hyperbolic manifold, which, like the $d$-spheres, is a $d$-manifold embedded in $\R^{d+1}$, defined as 
\begin{align}
    \mathbb{H}^{d}_K := \{x = (x_0,\dots,x_d) \in \mathbb{R}^{d+1}:  \langle x, x \rangle_{\mathcal{L}} = 1/K, \ x_0 > 0 \},
\end{align}
where $K<0$ is the curvature of the manifold, and $\langle \cdot, \cdot \rangle_{\mathcal{L}}$ is the Lorentzian inner product
\begin{align}
    \langle x, y \rangle_{\mathcal{L}} = -x_0y_0 + x_1y_1 + \dots + x_ny_n.
\end{align}
In our experiments, $K=-1$.

The $d+1$-dimensional Euclidean space endowned with the Lorentzian inner product $(\R^{d+1}, \langle \cdot,\cdot  \rangle_{\mathcal{L}} )$ is known as the Minkowski space. 
The Lorentz inner product is in general indefinite. Therefore, technically it is not an inner product. 
But it is positive definite when restricted to $\sH^d_K$, and as a result induces a valid Riemannian metric $g_\gL$.
Equation (\ref{eq:riem-div-ambient}), however, relies on the Euclidean geometry of the ambient space. 
Therefore, we model the density $p_\gE$ associated with the metric tensor $g_\gE$ induced by the regular Euclidean inner product. 
That is, $p_\gE$ is a probability density of the manifold $(\sH^d_K, g_\gE)$.
Note that all the data points still lie on the same topological space $\sH^d_K$, and the density can be translated via $p_\gE = p_\gL \sqrt{\frac{|G_\gL|}{|G_\gE|}}$, where $G_\gL$ and $G_\gE$ are the components of the matrix $g_\gL$ and $g_\gE$, and $p_\gL$ is the actual density on the Hyperbolic manifold $(\sH^d_K, g_\gL)$. 
This change-of-volume relation implies instead of maximizing the likelihood $\log p_\gL$, we can simply maximize $\log p_\gE$.

Alternatively, one can also compute the Riemannian divergence wrt the metric $g_\gL$ using the internal coordinates, as is done in \citep{lou2020neural}. 
In this case, the learned density will be the actual density $p_\gL$ on the hyperbolic manifold.

\paragraph{Tangential projection}
Similar to the spheres, we analyze the contribution of the differential $\ddd x$. 
\begin{align}
    \ddd \langle x, x\rangle_{\mathcal{L}} = 2 n_x \, \ddd x = 0,
\end{align}
where $n_{x} = (-x_0, x_1, \dots, x_d)$ is the normal vector. 
Subtracting the normal contribution gives rise to the tangential projection $P_x= I - \frac{n_xn_x^\top}{\norm{n_x}_2^2}$. 

Note that this is different from the usual ``Lorentz'' orthogonal projection $P^\gL_x(u) = u - \frac{\langle x, u\rangle_{\mathcal{L}}}{\langle x, x\rangle_{\mathcal{L}}} x $ \citep{Ratcliffe94}; the latter is not orthogonal in the Euclidean inner product.

\begin{figure}
    \centering
    \includegraphics[width=0.4\textwidth]{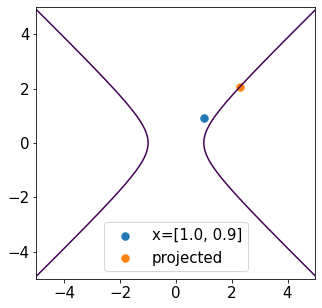}
    \caption{Closest-point projection of the point $(1.0, 0.9)$ onto the Hyperbolic manifold $\sH^1_{-1}$ in the Lorentz norm. This projection is clearly not the closest one in Euclidean distance. }
    \label{fig:hyperbolic_projection}
\end{figure}

\paragraph{Closest-point projection}
We first derive the closest-point projection wrt the Lorentz inner product. 
For any $x\in \{x': \langle x', x'\rangle_{\mathcal{L}} < 0 \}$, 
\begin{align}
    \pi(x) = \argmin_{y\in \sH^d_K} \norm{x-y}_\gL^2,
    \label{eq:closest-lorentz-objective}
\end{align}
where $\norm{x}_\gL:=\sqrt{\langle x, x\rangle_{\mathcal{L}} }$ is the Lorentz norm. 
To deal with the constraint $y\in\sH^d_K$, we can introduce the Lagrange multiplier $\lambda$, and find the stationary point of the function
\begin{align}
    \norm{x-y}_\gL^2 + \lambda (\langle y, y \rangle_{\mathcal{L}} - 1/K).
\end{align}
Taking the gradient wrt $y$ and setting it to be zero yield
\begin{align}
    -2n_{x-y} + 2\lambda n_y = 0 \quad\Longleftrightarrow \quad y = \frac{1}{\lambda + 1} x.
\end{align}
On the other hand, $y\in \sH^d_K$, which means $\lambda+1 = \sqrt{K\norm{x}_\gL}$, and therefore
\begin{align}
    \pi(x) = \frac{x}{\sqrt{K\norm{x}_\gL}}.
\end{align}
This projection, however, is not the closest-point projection in Euclidean distance in general, as depicted in Figure~\ref{fig:hyperbolic_projection}.
This is contrary to the claim made by \citet{skopek2019mixed}. 
In fact, following the same derivation (using Euclidean distance in place of the Lorentz norm in (\ref{eq:closest-lorentz-objective})), we would end up with a Lagrange multiplier that cannot be analytically solved, as it involves solving a root finding problem.

This projection, albeit not the shortest one in Euclidean distance, is still a valid projection. 
We use it in numerical integration to simulate the dynamics.

\subsection{Orthogonal groups}
The orthogonal groups are defined as $O(n)=\{X\in \R^{n\times n}: X^\top X = X X^\top = I\}$.
The determinant of $X$ is either $1$ or $-1$. 
The subgroup with determinant $1$ is called the special orthogonal group, denoted by $SO(n)$. 
Naturally, $\R^{n\times n}$ is an ambient space of the orthogonal groups.

\paragraph{Tangential projection}
Following the differential analysis, 
\begin{align}
    \ddd (XX^\top) &= X \ddd X^\top + \ddd X X^\top  =0.
\end{align}
That is, $\ddd X X^\top$ is  skew-symmetric.
Denote the set of skew-symmetric matrices by
$\Skew_n=\{X\in \R^{n\times n}: X^\top = -X\}$.

Let $U$ be an arbitrary matrix in $\R^{n\times n}$. We want to project it orthogonally onto $\gT_X O(n)$. 
The orthogonal projection needs to be the closest-point projection onto the subspace. 
We can use the Frobenius norm to induce the Euclidean distance metric over the entries of the matrix.
Then finding the closest-point projection $V$ of $U$ amounts to finding the stationary point of
\begin{align}
    \norm{U - V}_F^2 + \langle \Lambda, X V^\top + V X^\top\rangle_F,
\end{align}
where $\Lambda$ is the Lagrange multiplier.
Taking the gradient wrt $V$ yields
\begin{align*}
    \frac{\ddd}{\ddd V}  \langle U-V , U-V\rangle_F + \langle \Lambda, XV^\top + V X^\top\rangle_F &=\frac{\ddd}{\ddd V} \tr((U-V)^\top (U-V) + \Lambda^\top(XV^\top + VX^\top)) \\
    &= \frac{\ddd}{\ddd V} \tr(- 2U^\top V + V^\top V + VX^\top  \Lambda + XV^\top  \Lambda )\\
    &= - 2U + 2 V + \Lambda^\top X + \Lambda X. 
\end{align*}
Equating the last step with $0$ yields
\begin{align}
    V = U + \frac{\Lambda+\Lambda^\top}{2} X.
    \label{eq:orthogonal_group_tangential_lagrange}
\end{align}

Since $V$ needs to satisfy $XV^\top + VX^\top=0$, 
we have
\begin{align}
    XV^\top + VX^\top &= XU^\top + \frac{\Lambda + \Lambda^\top}{2} + UX^\top + \frac{\Lambda+\Lambda^\top}{2} \\
    &= XU^\top +  UX^\top + \Lambda+ \Lambda^\top = 0,
\end{align}
which means
\begin{align}
    \Lambda+ \Lambda^\top = -XU^\top -  UX^\top.
\end{align}

Substituting this into (\ref{eq:orthogonal_group_tangential_lagrange}) yields
\begin{align}
    V = \frac{ U -XU^\top X }{2}.
\end{align}

That is, $P_X(U) = \frac{ U -XU^\top X }{2} $ for orthogonal groups.

\paragraph{Closest-point projection}
Again, using the Lagrange multiplier $\Lambda$ for the constraint that the projection $M$ of $X$ should satisfy $M^\top M=I$, we try to find the stationary point of the following quantity
\begin{align}
    \norm{M - X}_F^2 + \langle \Lambda, M^\top M - I\rangle_F.
\end{align}

Equating the gradient wrt $M$ with $0$ gives
\begin{align*}
    \frac{\ddd}{\ddd M} \langle M-X, M-X\rangle_F + \langle \Lambda, M^\top M - I\rangle_F 
    &= \frac{\ddd}{\ddd M} \tr( (M-X)^\top (M-X) +(M^\top M - I)^\top\Lambda) \\
    &= \frac{\ddd}{\ddd M} \tr( M^\top M - 2X^\top M + M^\top M \Lambda) \\
    &= 2M - 2X + M\Lambda + M \Lambda^\top = 0,
\end{align*}
which means
\begin{align}
    M = 2X ( 2 I + \Lambda + \Lambda^\top)^{-1}.
    \label{eq:ortho_group_projection_lagrange}
\end{align}

Since $M$ is orthogonal, we have
\begin{align}
     M^\top M = 4( 2 I + \Lambda + \Lambda^\top)^{-T}  X^\top X ( 2 I + \Lambda + \Lambda^\top)^{-1}  = I,
\end{align}
which means
\begin{align}
    4 X^\top X = ( 2 I + \Lambda + \Lambda^\top)^2.
\end{align}

Let $X=UDV^\top$ be the singular value decomposition of $X$. 
Then 
\begin{align}
    2 V D V^\top  =  2I + \Lambda + \Lambda^\top.
\end{align}

Substituting this into (\ref{eq:ortho_group_projection_lagrange}), we get
\begin{align}
    M = X  V D^{-1} V^\top = UDV^\top V D^{-1} V^\top = U V^\top.
\end{align}
That is, $\pi(X)=UV^\top$ for orthogonal groups, where $U,V$ are the left and right singular matrices of $X$.

\section{Experimental details}
\label{sec:experimental-details}

\subsection{Architecture}
In our experiments, we parameterize the $a$ network as a multi-layer perceptron (MLP) with either the sinusoidal or the swish activation function. 
For the hyperbolic experiments, the first layer of the MLP has an additional ActNorm layer \citep{kingma2018glow} which we find adds extra numerical stability. The ActNorm layer is initialized before training with one batch such that its output has a mean of zero and a standard deviation of one. In an analogous manner to training the MLP the ActNorm parameters are updated via backpropagation. For the orthogonal group experiments we flatten the input matrix into a vector before passing it to the MLP. The details of our various model are given in Table \ref{tab:arch-detail}. For our importance sampler which is used to represent a differentiable distribution over $[0, T]$, we use a deep sigmoidal flow \citep{huang2018neural} (without the final logit activation) followed by a fixed scaling flow, which represents the range $[0,T]$.
We disconnect the gradient from the numerical solver to save compute; \ie $Y_s$ is not differentiable.
This would result in slightly biased gradient updates for minimizing the variance of the importance estimator, but we still observe substantial reduction in variance (see Figure~\ref{fig:imp_ablation}). 
Finally, we use \textit{PyTorch} \citep{paszke2019pytorch} as our deep learning framework.

\xhdr{Computational Resources}
We run all of our experiments either on a single NVIDIA Tesla V100 or a single NVIDIA Quadro RTX $8000$ GPU for a maximum of $30$ hours.

\begin{table}[t]
    \small
    \centering
    \begin{tabular}{l|cccc}
    \textbf{Manifold} & \textbf{Activation} & \textbf{Hidden layers} & \textbf{Embedding size} & \textbf{ActNorm first}\\
    \midrule
    Sphere & Sine & 5 & 512 & False \\
    Tori & Swish & 4 & 256 & False\\
    Hyperbolic & Swish & 2 & 512 & True\\
    Orthogonal group & Swish & 256 & 256 & False\\
    \bottomrule
    \end{tabular}
    \caption{ 
    \small
    The variational function $a$ network architectures for different manifolds in our experiments.
    }
    \label{tab:arch-detail}
\end{table}

\subsection{Optimization}
We use the Adam \citep{kingma2014adam} optimizer to train the $a$ network. The learning rate and momentum parameters used for each manifold is mentioned in the Table \ref{tab:exp-optimization-detail}. 
For the sphere experiments, we slowly decrease the learning rate during training using a cosine scheduler.
For optimization of our importance sampler, we use Adam with a fixed learning rate of $0.01$. 
We update the importance sampler every $500$ steps of our training loop for the $a$ network.
Lastly, to optimize our mixture of power spherical distributions for the tori experiments we use Adam with a learning rate of $0.03$ with $\beta_{1}=0.9$ and $\beta_{2}=0.999$. 

\begin{table}[t]
    \small
    \centering
    \begin{tabular}{l|ccccc}
    \textbf{Manifold} &  \textbf{Optimizer} &\textbf{Learning rate} & $\mathbf{\beta_{1}}$ & $\mathbf{\beta_{2}}$ & \textbf{Scheduler}\\
    \midrule
    Sphere & Adam & $2e-4$ & $0.9$ & $0.999$ & Cosine \\
    Tori & Adam & $3e-4$ & $0.9$ & $0.999$ & None \\
    Hyperbolic & Adam & $5e-4$ & $0.9$ & $0.999$ & None \\
    Orthogonal group & Adam & $1e-3$ & $0.9$ & $0.999$ & None \\
    \bottomrule
    \end{tabular}
    \caption{\small Optimization hyperparameters for experiments on different manifolds}
    \label{tab:exp-optimization-detail}
\end{table}

\subsection{KELBO}
The gap between the exact likelihood of the data given the model, \ie $\log p(x)$, and the Riemannian CT-ELBO may be large. This evaluation gap makes empirical validation of the models using the Riemannian CT-ELBO imprecise. We acquire a tighter lower bound by using $K > 1$ samples and importance sampling similar to \citet{burda2015importance}. In details, we know from (\ref{eq:exactlogp_in_thm2}) that:
\begin{align*}
& \log p(x, t)=\log \mathbb{E}_\sQ\left[ \frac{\ddd\sP}{\ddd\sQ}\cdot p_0\left(Y_{t}\right) \exp \left(- \int_{0}^{T} 
\nabla_g\cdot \left(V_0 - \frac{1}{2}  (V \cdot \nabla_g) V\right)
\ds\right) \, \Bigg\vert \, \vy_0=x \right].
\end{align*}
We rewrite this as:
\begin{align*}
& \log p(x, t)=\log \mathbb{E}_\sQ L(\vy),
\end{align*}
where $L(\vy)$ is defined to be:
\begin{align*}
\frac{\ddd\sP}{\ddd\sQ}\cdot p_0\left(Y_{t}\right) \exp \left(- \int_{0}^{T} 
\nabla_g\cdot \left(V_0 - \frac{1}{2}  (V \cdot \nabla_g) V\right)
\ds \right).
\end{align*}
Then by Jensen's inequality, we have that:
\begin{align*}
& \log p(x, t)=\log \mathbb{E}_\sQ L(\vy)
= \log \mathbb{E}_\sQ  \sum_{i=1}^{K} \frac{1}{K} L(\vy^{i}) \geq  \mathbb{E}_\sQ \log \sum_{i=1}^{K} \frac{1}{K} L(\vy^{i}),
\end{align*}
where $Y^{i}$s are \iid trajectories sampled from $\sQ$. We call this new lower bound KELBO.
Note that this is a tighter lower bound because we can write:
\begin{align*}
    & \text{KELBO} = \mathbb{E}_\sQ \log \sum_{i=1}^{K} \frac{1}{K}  L(\vy^{i})
    \geq \mathbb{E}_\sQ  \sum_{i=1}^{K} \frac{1}{K} \log L(\vy^{i}) = 
    \mathbb{E}_\sQ  \log  L(\vy) = \text{Riemannian CT-ELBO}.
\end{align*}
In fact, this lower bound increases monotonically to the true likelihood as $K\rightarrow\infty$.
We use KELBO with $K=100$ for evaluating all of our models. We have experimented with the $K$ to be up to $1000$ and found out the results stop changing much for $K>100$.
\label{sec:kelbo}

\subsection{Numerical integration of the SDEs}

During training and evaluation, we numerically integrate the SDE on each respective manifold using the Stratonovich-Heun method as described in \citet{burrage2004numerical}. 
Each iteration is followed by the closest-point projection (in the case of $\sH^d_K$, we use the closest-point project wrt the Lorentz inner product).
The number of integration steps for each manifold during training is reported in Table \ref{tab:elbo_steps}.

\begin{table}[t]
    \small
    \centering
    \begin{tabular}{lc}
    \toprule
    \textbf{Manifold} & \textbf{Integration steps during training} \\
    \midrule
    Sphere & 100\\
    Tori & 1000 \\
    Hyperbolic & 100\\
    Orthogonal group & 100\\
    \bottomrule
    \end{tabular}
    \caption{
    \small
    Details of training integration.
    }
    \label{tab:elbo_steps}
\end{table}

During evaluation, as described in \ref{sec:kelbo}, we numerically integrate the data from $s=0$ to $s=T$, and the It\^o integral involved in the KELBO is approximated using the Euler-Maruyama scheme (note that the dynamics is still generated using Stratonovich-Heun).
As computing KELBO requires forward passes through the $a$ network, it may not be as smooth as just integrating the inference SDE. 
Therefore, we use an adaptive step size for integration.
We adapted the \textit{torchsde} library \citep{kidger2021neuralsde,li2020scalable} to calculate errors and adapt the step size accordingly. 
The error tolerance and minimum step size used in integration for all the experiments are reported in Table \ref{tab:adaptive-detail}. 
Also, for plotting densities we use the exact log likelihood of the equivalent ODE. 
To numerically integrate the ODE for computing the exact likelihood, we use the default \texttt{dopri5} solver from the \textit{torchdiffeq} library \citep{chen2018neuralode,chen2021eventfn}. 
Finally, we use 
\textit{cartopy} \citep{Cartopy},
\textit{matplotlib} \citep{hunter2007matplotlib},
and \textit{plotly} \citep{plotly} for visualization. 

\begin{table}[t]
    \small
    \centering
    \begin{tabular}{ccc}
    \toprule
    \textbf{rtol} & \textbf{atol} & \textbf{minimum step size} \\
    \midrule
    $1e-3$ & $1e-3$ & $1e-5$ \\
    \bottomrule
    \end{tabular}
    \caption{Configuration of the adaptive step size integration used during evaluation.}
    \label{tab:adaptive-detail}
\end{table}

\end{document}